\documentclass[11pt,a4paper]{article}

\usepackage[T1]{fontenc}
\usepackage[utf8]{inputenc}
\usepackage{lmodern}
\usepackage{microtype}
\usepackage[margin=1in]{geometry}
\usepackage{amsmath,amssymb}
\usepackage{graphicx}
\usepackage{booktabs}
\usepackage{tabularx}
\usepackage{array}
\usepackage{multirow}
\usepackage{caption}
\usepackage{subcaption}
\usepackage{enumitem}
\usepackage{xcolor}
\usepackage{textcomp}
\usepackage{listings}
\usepackage[round,authoryear]{natbib}
\usepackage[colorlinks=true,linkcolor=blue!60!black,citecolor=blue!60!black,urlcolor=blue!60!black]{hyperref}
\usepackage{url}
\makeatletter\g@addto@macro\UrlBreaks{\do\-\do\_}\makeatother

\AtBeginDocument{}

\title{The Linguistic Knowledge Benchmark: Do Language Models Know the Foundations of Early-Grade Literacy?}

\author{
  Maxime Leli\`evre \qquad
  Alasdair Mackintosh \qquad
  Dominic Andrews \\[0.4em]
  Tobias Murray \qquad
  Moritz Boos \qquad
  Oliver G.\,B. Garrod \\[0.8em]
  Fab AI
}

\date{}

\begin{document}

\maketitle

\begin{abstract}
Many languages of instruction in low- and middle-income countries (LMICs) are poorly represented in the data used to train language models. Before these models can support early-grade literacy in such languages, we need to know whether they hold the linguistic knowledge of each language that literacy teaching relies on. To test this, we present the Linguistic Knowledge Benchmark and the reproducible, largely automated pipeline that turns curriculum documents, teacher guides and pupil books into a vetted question set organised on a fixed grid of nine linguistic domains and seven grade levels. The pipeline combines LLM-driven source survey, feature consolidation, item generation and quality control with an item-selection step inspired by item response theory and a final expert review. We instantiate it for Kiswahili (Kenya; 135 items) and English (United Kingdom; 137 items) and evaluate 121 and 119 models respectively. Frontier models exceed 90\% accuracy in both languages, but accuracy falls steeply for smaller and cheaper models, and more so in Kiswahili. The two best models under 10 billion parameters reach 91.7\% and 73.0\% in English but only 51.9\% and 40.7\% in Kiswahili, and the mean accuracy of models deployable on mobile and tablet hardware is 47\% in English against 24\% in Kiswahili. The models most practical for low-resource classrooms are therefore the least capable in the languages they would serve. We release a 20-item sample per language, all pipeline prompts, open-source code, and a public leaderboard. This is the first rung of a three-level evaluation ladder for measuring and generating decodable text to support foundational literacy. 
\end{abstract}

\section{Introduction}
\label{sec:intro}

Every language is different. In Kiswahili, grammatical number (singular/plural) is often marked through changes to the start of nouns, whereas in English plurals are usually marked with changes to the end of nouns. Pronunciation, word order and tense all vary across languages. How well language models understand and generate text in a language depends in part on how much exposure, and what kind of exposure, they have had to that language.

Many languages used in LMICs are poorly represented in AI training data, giving language models fewer opportunities to learn their linguistic structures and patterns. Common Crawl, a public archive of web pages widely used to train AI models, shows this imbalance: African languages account for well below one per cent of its identified content \citep{commoncrawl2026}.

If AI is going to help solve the learning crisis, developers and users need to know whether it works in the languages used in LMIC contexts, where mother-tongue instruction is known to have large benefits for foundational literacy \citep{geeap2025reading}. Much of what teachers need in early-grade literacy is level-specific: practice readers, targeted support and assessment passages, each restricted to the letter--sounds already taught. Such material is called \emph{decodable text}. Decodable material is scarce in Kiswahili and many other LMIC languages, and AI could help close that gap.

Our end goal is to evaluate whether AI models can generate decodable text. A single end-to-end score would show whether models succeed, but not why they fail, and therefore not where to intervene. We therefore built a three-rung benchmark ladder that separates the capabilities decodable-text generation depends on. \emph{Linguistic Knowledge} tests what a model knows about a language's key features. \emph{Decodability Assessment} tests whether a model can identify words a child cannot yet decode, given a specified instructional sequence. \emph{Decodable Text Generation} tests whether a model can write text that is decodable. This paper concerns the first rung.

Our contributions are as follows:
\begin{itemize}[leftmargin=1.5em,itemsep=2pt]
  \item A reproducible, language-agnostic pipeline that turns linguistic knowledge gathered from primary sources into a vetted evaluation benchmark, with automated stages for source survey, consolidation, item generation and quality control, an item-selection step inspired by item response theory, and a final expert review (Section~\ref{sec:methods}).
  \item Two instantiations of the pipeline: a Kiswahili benchmark built from Kenya-focused resources (135 items) and an English benchmark built from UK-focused resources (137 items), each validated at the item-selection, expert-review and scoring levels.
  \item An evaluation of 121 models in Kiswahili and 119 in English, showing that frontier models perform very well in both languages while the smaller models that are more practical for on-device and resource-constrained deployment struggle, with a consistently larger deficit in Kiswahili (Section~\ref{sec:results}).
\end{itemize}

\section{Methods}
\label{sec:methods}

\subsection{Pipeline}

We designed the Linguistic Knowledge Benchmark as a reproducible pipeline so that it can be adapted and produced for additional languages. The pipeline turns source documents (curricula, reference grammars, teacher guides and pupil books) into a vetted question set that assesses how well a model can answer questions about language features relevant to foundational literacy. Every question sits on a fixed grid of nine linguistic domains and school grade levels (from pre-primary to Grade~5 and beyond) that stays the same across languages (Table~\ref{tab:grid}). The grade tag records where the content is \emph{first introduced} in the curriculum, not how hard the question is. Each item also carries two framework tags: the strand of the Simple View of Reading (SVR; \citealp{gough1986decoding}) that the tested knowledge serves (decoding or language comprehension), and the core instructional component identified by the Global Education Evidence Advisory Panel (GEEAP; \citealp{geeap2025reading}) that an error in the item would corrupt (Table~\ref{tab:schema}). The methodology and linguistic domains were reviewed with sector pedagogy experts.

Importantly, the benchmark targets \emph{formal} linguistic competence (getting the form of a language right) rather than \emph{functional} competence (using the language to accomplish goals in the world), the two being distinct cognitive skills \citep{mahowald2024dissociating}. The two upper rungs, Decodability Assessment and Decodable Text Generation, test functional competence and are currently under development.

\subsection{Design rationale}

The pipeline was built following three guiding principles.
\begin{itemize}[leftmargin=1.5em,itemsep=2pt]
  \item \textbf{Mirror the expert workflow.} The pipeline keeps the main stages of benchmark construction separate: reviewing the source documents, inventorying testable facts, consolidating them into a unified map of the language, drafting questions against that map, quality-checking them, and selecting a balanced final set. Keeping each stage separate creates a traceable process, making it easier for experts to identify where quality degrades.
  \item \textbf{Use experts selectively.} The pipeline automates source-document analysis, extraction, question drafting and initial quality control, and uses domain experts for the final review. This reduces the amount of expert time required per benchmark while retaining human judgement.
  \item \textbf{Reuse across languages.} Every stage uses a language-agnostic prompt, so adapting the benchmark to a new language only requires swapping the source documents and revising the language-specific content of the prompts (Table~\ref{tab:adaptation}).
\end{itemize}

\subsection{Orchestration}

\begin{figure}[t]
  \centering
  \includegraphics[width=0.9\linewidth]{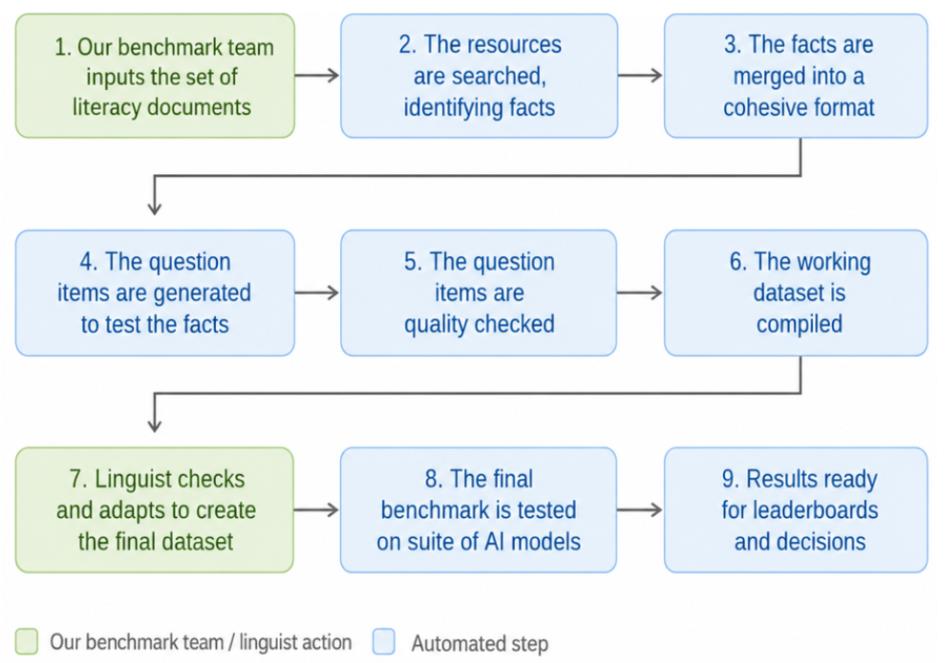}
  \caption{End-to-end pipeline for producing a Linguistic Knowledge Benchmark in a target language. Green indicates steps performed by the benchmark team or a linguist; blue indicates automated steps.}
  \label{fig:pipeline}
\end{figure}

The benchmark development team inputs a curated set of resources (typically 6 to 11) covering the standard early-literacy topics in the target language: curriculum documents, teacher guides and pupil-facing books. The resources used for each language are listed in Tables~\ref{tab:sources-sw} and~\ref{tab:sources-en}. The pipeline classifies each source into one of four tiers: curriculum, teacher material, student material, and other (e.g.\ a grammar book). Figure~\ref{fig:pipeline} summarises the end-to-end process. The pipeline then proceeds as follows.

\begin{enumerate}[leftmargin=1.5em,itemsep=3pt]
  \item \textbf{Survey the resources} (Prompt~\ref{prompt:stage1}). One document at a time, the pipeline compiles a list of linguistic facts. For each fact it records a verbatim example, a document location (file, page, heading), the domain, and the grade at which the content is first introduced. Working per document means sources can be processed in parallel or added later.
  \item \textbf{Merge the extracted facts} (Prompt~\ref{prompt:stage2}). All facts are consolidated into a single feature map, grouped by domain and ordered by grade. Duplicates are merged into one feature listing all of its contributing sources. Where two sources disagree on a grade, the highest-tier source wins, according to a manual hierarchy given in the prompt.
  \item \textbf{Generate question items} (Prompt~\ref{prompt:stage3}). Items are either multiple choice or short answer, depending on the nature of the fact: contrasts and judgements are tested through multiple-choice items, while facts about producing a form use short-answer items. Using both formats lets us test unaided recall as well as the ability to distinguish correct forms from plausible, diagnostic errors. This stage is run once per subset of sources small enough to fit in the context window. It deliberately oversamples, generating several candidates per feature and making no global decisions about counts or balance.
  \item \textbf{Quality-check the items} (Prompt~\ref{prompt:stage4}). All candidates are pooled and each is checked against explicit anti-gaming and shape rules. For instance, the answer must not be given away in the stem, options must be length-parallel and functionally plausible, and the item must not be solvable by general reasoning alone. Duplicates are grouped and ranked rather than deleted, so nothing is discarded without a record.
  \item \textbf{Compile a working dataset.} Items are selected using an approach inspired by item response theory (IRT; \citealp{lord1980irt}), a statistical framework for judging how well each question separates stronger from weaker performers. Using the responses of a panel of 10 models spanning different providers and a wide performance range (Table~\ref{tab:panel}), we select a trimmed set that (a)~avoids saturation by dropping items every model gets right, (b)~maximises differentiation by favouring high-discrimination items, and (c)~keeps a balanced domain $\times$ grade spread. Items that no model answers correctly are deliberately kept and flagged for the human expert to review.
\end{enumerate}

A linguist or first-language speaker then reviews every question, adapting or discarding any items with remaining issues to create the final benchmark dataset. The automated process tracks sources throughout, which keeps the build verifiable, focuses human expertise on quality review, and keeps the overall process less costly. After the final human review, we check that the correct answers to multiple-choice items are evenly distributed across the four options, so that position is not a usable cue.

Every item asks about the target language but is written in English. This keeps what we are measuring separate from what we are not: asking about a language's structure in that language confounds linguistic knowledge with reading comprehension and instruction-following in the same language, so a wrong answer is hard to attribute. Posing questions in English isolates whether the model knows the formal properties of the target language, and matches some real deployments, where a tool may be operated in English while producing local-language content. This is a favourable setting, so results should be read as an upper bound (Section~\ref{sec:limitations}). It also leaves open the option of translating the items later. Table~\ref{tab:examples} shows two items from the public Kiswahili sample, one in each answer format.

\begin{table}[t]
  \centering
  \small
  \begin{tabularx}{\linewidth}{@{}>{\raggedright\arraybackslash}p{4.2cm}>{\raggedright\arraybackslash}X@{}}
    \toprule
    Domain (grade) & Item \\
    \midrule
    Phonics and orthography (Grade~1)\newline\textit{multiple choice} &
    In the written word \textit{fedha}, the letters \textit{d} and \textit{h} together stand for how many consonant sounds?\newline
    (A)~one single sound\newline (B)~three separate sounds\newline (C)~two separate sounds\newline (D)~no sound at all\newline
    \textit{Correct answer:} A \\
    \addlinespace
    Metalinguistic reasoning (Grade~3)\newline\textit{short answer} &
    What is the Swahili term for the negated form of a verb (the negation)?\newline
    \textit{Correct answer:} \textit{ukanusho} \\
    \bottomrule
  \end{tabularx}
  \caption{Two items from the public Kiswahili sample. Both are written in English and ask about Kiswahili. The grade is where the content is first introduced in the Kenyan curriculum, not the difficulty of the item.}
  \label{tab:examples}
\end{table}

All prompts are reproduced in Appendix~\ref{app:prompts}. The code is open source on GitHub\footnote{\label{fn:github}\url{https://github.com/Fab-Inc/qa-benchmarks-monorepo}} as a package that can be run via the APIs of several providers.

\subsection{Pipeline validation}
\label{sec:pipeline-validation}

We validated the pipeline at two levels: the effect of item selection and the outcome of expert review.

\begin{figure}[t]
  \centering
  \includegraphics[width=\linewidth]{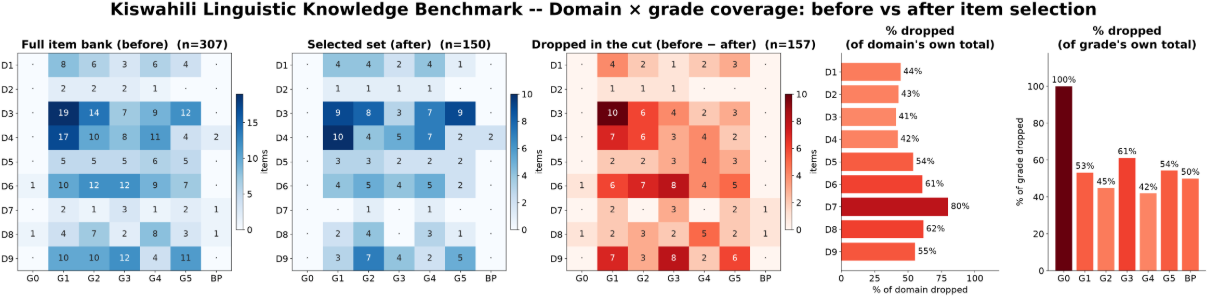}
  \caption{Kiswahili item coverage across linguistic domains (D1--D9) and grade levels (G0--G5, BP = beyond primary) before and after item selection. See Figure~\ref{fig:coverage-en} for English.}
  \label{fig:coverage-sw}
\end{figure}

\paragraph{Item selection.}
For Kiswahili, item selection reduces the pool from 307 to 150 items, meeting the target of 150 set in the selection code. Of the 157 items removed, 85 were answered correctly by all ten models and therefore carry no discriminating power. The remainder were trimmed on difficulty, discrimination, and spread across linguistic domains and grade levels. For English, the same procedure reduces 377 items to 140, marginally below target, as 217 items were answered correctly by every model. Figure~\ref{fig:coverage-sw} reports domain $\times$ grade coverage for Kiswahili before and after selection, together with the proportion of items dropped within each domain and each grade.

The reduction is broadly uniform: most domains and grades lose between 40\% and 60\% of their items, so no single stratum absorbs a disproportionate share of the cut. Two strata depart from this pattern, both of which were sparsely populated in the full bank: D7 (Pragmatics and Discourse), which loses 80\% of its items, and Grade~0, which is dropped in full and is therefore not represented in the released set. All remaining domains and grades retain coverage. The corresponding figure for English is given in Figure~\ref{fig:coverage-en}. In both languages, model ranking by accuracy is unchanged between the full pool and the selected set (Spearman $\rho = 1.00$, Kendall $\tau = 1.00$), indicating that the selected set induces the same ordering over models as the full pool. Table~\ref{tab:selection} reports the selection counts and both validation diagnostics for each language. To check that the selected set does not depend on the particular composition of the panel, we also re-ran the selection once for every pair of panel models left out (45 runs) and recorded, for each item in the selected set, the share of runs in which it was selected again. The mean of this selection frequency is 0.90 for English and 0.89 for Kiswahili, so the selected items are largely stable under changes to the panel.

\paragraph{Expert review.}
Every candidate item passing the selection step was read by a human expert before entering the final set. Reviewers worked from a fixed rubric separating two judgements per item: whether the item itself is sound (accept as is, edit, or reject when the flaw was too large to repair), and, independently, whether its metadata is correct (domain, grade, SVR strand, GEEAP component; Table~\ref{tab:schema}). Kiswahili was reviewed twice, sequentially. A pedagogy expert reviewed all 150 candidates, dropping 12 and correcting 9. The resulting 138 items then went to an early-literacy expert and native Kiswahili speaker, who dropped a further 3 and edited 10, yielding the final 135. English was reviewed once, by a pedagogy expert and native English speaker, over 140 candidates: 3 were dropped and 7 rewritten, yielding the final 137.

Taken together, the experts' decisions to edit or drop items followed recurring patterns. Rejections traced to problems with how the items were generated, and the same problems came up in both languages. Redundancy dominated in Kiswahili, where most first-pass drops duplicated a linguistic feature covered elsewhere; answer leakage (the target answer recoverable from the stem) accounted for a smaller share. The remainder in both languages were items with more than one defensible answer, and items that measured the wrong thing (testing linguistics expertise rather than teacher knowledge, or covering content not taught in schools). Edits diverged by language. English corrections were almost entirely to metadata, most often reassigning an SVR strand or GEEAP component. Kiswahili second-pass edits mainly expanded an item's set of accepted answers (\texttt{expected\_answers}, Table~\ref{tab:schema}) with valid responses the generation stage had missed.

\subsection{Scoring}
\label{sec:scoring}

We use regular-expression parsing for multiple-choice items and LLM-as-a-judge for short-answer items. The prompts used for question answering and judging are reproduced in Prompts~\ref{prompt:answer} and~\ref{prompt:judge}.

For short answers, we run three judges, one each from three different providers, and take the majority verdict, to avoid the bias and prompt-sensitivity of any single judge \citep{verga2024juries}: Claude Sonnet~5 (Anthropic), GPT-5.6 Terra (OpenAI) and Gemini-3.6 Flash (Google). The judge prompt is strict: it accepts only differences of capitalisation, hyphenation and spacing, never synonyms or paraphrases (Prompt~\ref{prompt:judge}). Chain-of-thought blocks from a model's reasoning are stripped from the text sent to the judge, so that a long reasoning trace cannot dwarf the actual answer.

Each judge replies in a structured format with two fields: a boolean verdict and a written justification. The verdict is used directly to compute accuracy, while the justification serves two purposes. It requires the judge to ground its verdict in verifiable facts, and it makes it possible to inspect a judge's reasoning and disagreements between judges. For example, during development we noticed that some short-answer items were missing valid answers from their \texttt{expected\_answers} list (Table~\ref{tab:schema}), because many models consistently answered with a valid synonym that was not listed.

For multiple choice, six regular-expression patterns cover the common response formats, including answers wrapped in parentheses and answers that follow a closed reasoning block. When a model's response matches none of the patterns, we run a fallback pass that sends only those unparseable rows to a single LLM judge, asking whether the response indicates the correct letter (Prompt~\ref{prompt:judge}). Any multiple-choice answer that cannot be parsed is either resolved by the fallback judge or excluded from the accuracy denominator; it is never counted as wrong. Each scored row records which method produced its verdict (\texttt{regex}, \texttt{llm\_judge}, \texttt{llm\_fallback} or \texttt{unscored}), so that strict-format scores remain distinguishable from rescued ones. We report accuracy for each model computed over scored rows only, with the unscored count reported alongside.

\subsection{Scoring validation}
\label{sec:scoring-validation}

The scoring layer was validated alongside the items, with each check summarised below.

\paragraph{Judge agreement.}
Agreement between the three LLM judges on short-answer responses is reported as the raw pairwise rate and as Cohen's $\kappa$, which corrects for the inflation that raw agreement exhibits when most responses are correct. Pairwise agreement was higher in Kiswahili (97.5--98.7\%, $\kappa$ 0.944--0.970) than in English (92.8--96.4\%, $\kappa$ 0.773--0.897), with full three-way consensus on 97.0\% of judged short-answer responses in Kiswahili and 92.1\% in English. The remainder were 2-of-3 splits resolved by majority vote.

\paragraph{Self-preference bias.}
LLM judges can favour their own outputs \citep{panickssery2024self}, so we checked whether any judge favoured responses from models of its own provider. We compared each judge's leniency toward responses from its own provider with its leniency toward other providers' responses, where leniency denotes marking an item correct that the other two judges mark incorrect (rows where the other two disagree are excluded). We found no meaningful self-preference effect on scoring in either language.

\paragraph{Fallback rate.}
A single LLM judge (Claude Sonnet~5) re-scored multiple-choice rows where parsing failed, using Prompt~\ref{prompt:judge}. Fallback was never invoked for 90\% of models in English and 91\% in Kiswahili. Among models that required it, the higher-scoring models showed either an increase in accuracy or no change. Claude Sonnet~4.5 went from 92.2\% (95/103) to 94.2\% (129/137) in English, and from 78.2\% (79/101) to 81.5\% (110/135) in Kiswahili; Kimi~K2.5 went from 97.0\% (127/131) to 97.1\% (133/137) in English. Their recovered answers were correct about as often as the answers that parsed normally. The smallest models did not gain, and some lost accuracy: Qwen3-1.7B went from 53.6\% (52/97) to 48.9\% (67/137) in English, and Llama-3.2-1B from 27.3\% (36/132) to 27.0\% (37/137). Their recovered answers were mostly wrong.

\paragraph{Model-family consistency.}
We checked consistency across six model families (Claude, GPT, Gemini, Gemma, Llama, Qwen) in both languages. Accuracy respects model size and tier overall, with only isolated inversions among same-generation, same-tier variants (e.g.\ within GPT and Qwen). This is consistent with the benchmark measuring a capability that scales with model size (Figures~\ref{fig:family-en} and~\ref{fig:family-sw}).

\section{Results}
\label{sec:results}

\begin{figure}[p]
  \centering
  \includegraphics[width=\linewidth]{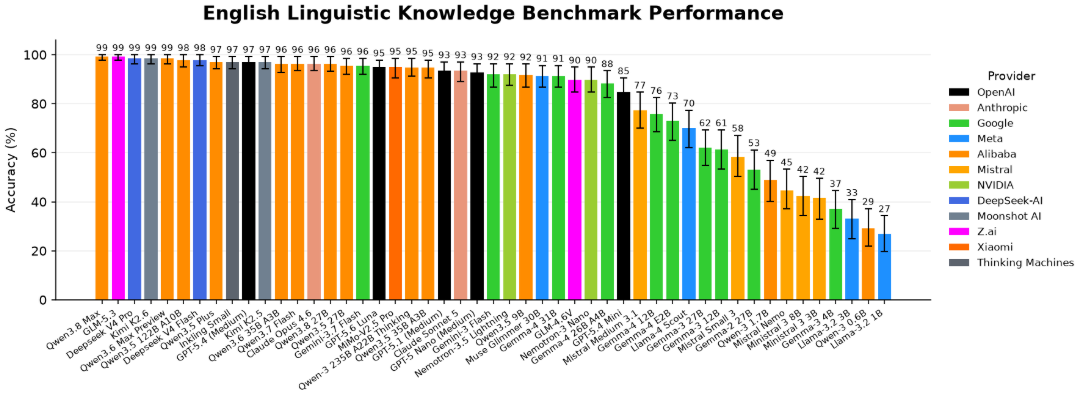}
  \caption{Accuracy on the English Linguistic Knowledge Benchmark ($n = 137$ items) for the 49 models pre-selected for the public leaderboard (of 119 evaluated). Accuracy is computed over scored rows only. Error bars show 95\% confidence intervals from 1,000 bootstrap replicates resampling items with replacement.}
  \label{fig:acc-en}

  \vspace{1.5em}

  \includegraphics[width=\linewidth]{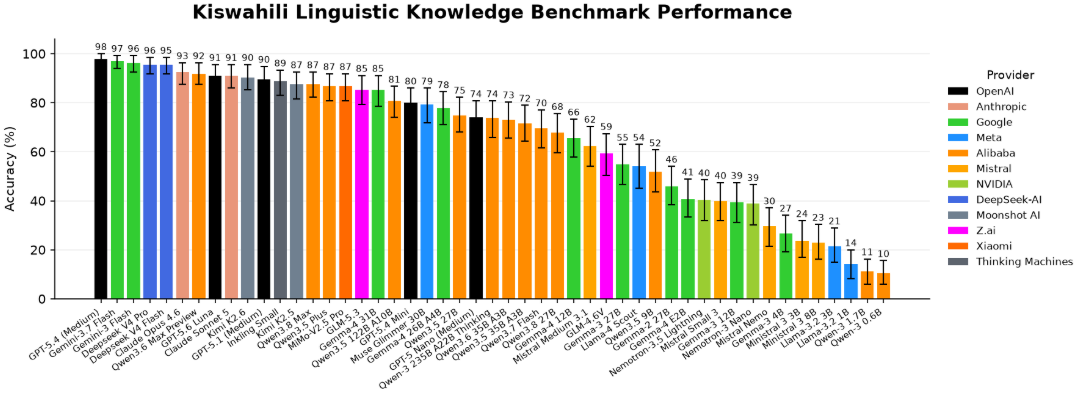}
  \caption{Accuracy on the Kiswahili Linguistic Knowledge Benchmark ($n = 135$ items) for the 49 models pre-selected for the public leaderboard (of 121 evaluated). Accuracy is computed over scored rows only. Error bars show 95\% confidence intervals from 1,000 bootstrap replicates resampling items with replacement.}
  \label{fig:acc-sw}
\end{figure}

Accuracy spans nearly the full available range in both languages, from 99\% for the highest-scoring model to 10\% for the lowest. Frontier models cluster above 90\%, while accuracy declines steeply among mid-sized and smaller models (Figures~\ref{fig:acc-en} and~\ref{fig:acc-sw}).

The two languages are not equivalent. English exhibits a longer plateau above 90\% and a shallower tail, whereas the Kiswahili distribution begins to fall away earlier and extends lower (Figure~\ref{fig:acc-sw}). Consequently, the set of models meeting any given accuracy threshold is substantially smaller in Kiswahili than in English. Figures~\ref{fig:acc-en} and~\ref{fig:acc-sw} show the subset of models pre-selected for the public leaderboard; accuracies for all evaluated models are reported in Tables~\ref{tab:all-en} and~\ref{tab:all-sw}.

\begin{figure}[t]
  \centering
  \begin{subfigure}{0.49\linewidth}
    \includegraphics[width=\linewidth]{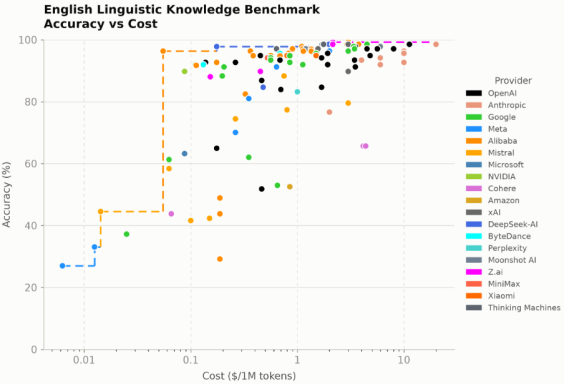}
    \caption{English, accuracy vs.\ cost}
  \end{subfigure}\hfill
  \begin{subfigure}{0.49\linewidth}
    \includegraphics[width=\linewidth]{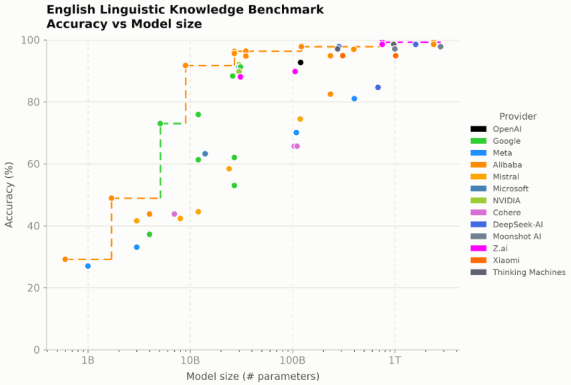}
    \caption{English, accuracy vs.\ model size}
  \end{subfigure}

  \vspace{0.5em}

  \begin{subfigure}{0.49\linewidth}
    \includegraphics[width=\linewidth]{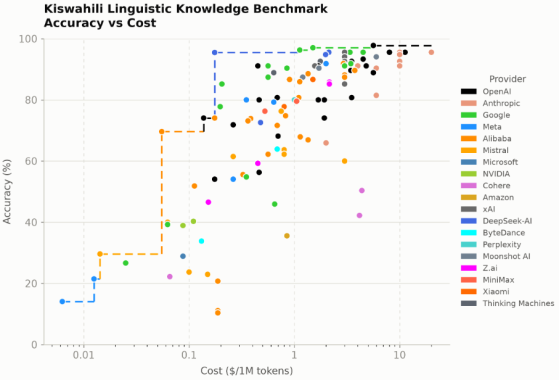}
    \caption{Kiswahili, accuracy vs.\ cost}
  \end{subfigure}\hfill
  \begin{subfigure}{0.49\linewidth}
    \includegraphics[width=\linewidth]{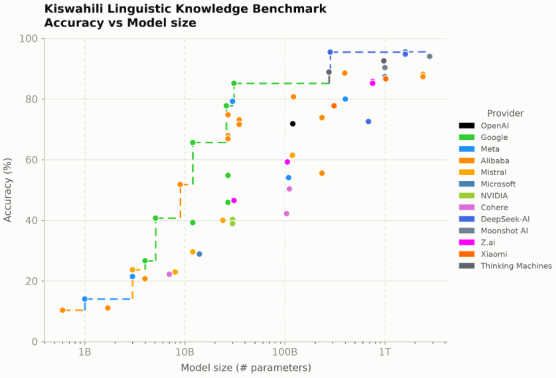}
    \caption{Kiswahili, accuracy vs.\ model size}
  \end{subfigure}
  \caption{Linguistic Knowledge accuracy against inference cost (left) and model size (right), for English (top) and Kiswahili (bottom). Each point is one model; colour denotes provider. Both x-axes are logarithmic. The dashed step line traces the Pareto frontier, i.e.\ the highest accuracy attainable at or below each cost or size.}
  \label{fig:pareto}
\end{figure}

These results have a direct consequence for deployment. The models that perform strongly on the benchmark are the largest and most costly to serve, whereas the affordable, offline-capable models (those that could plausibly operate in a low-resource classroom) are the least able to hold the linguistic knowledge a foundational-literacy application would depend on. At or below \$0.10 per million tokens (assuming a 3:1 ratio of input to output tokens), the best model available at the time of writing is Qwen3.7 Flash, which reaches 96.4\% in English but 69.6\% in Kiswahili. Among models small enough to run on a phone or tablet (below 10 billion parameters), the two best are Qwen3.5~9B and Gemma-4~E2B, at 91.7\% and 73.0\% in English but 51.9\% and 40.7\% in Kiswahili. Strong performance is therefore achievable, but not at the price point or under the operating constraints that make such tools practical to deploy, and the gap for smaller models is larger in Kiswahili than in English (Figure~\ref{fig:pareto}).

We grouped models by hardware tier so that accuracy could be compared across the classes of device available in LMIC classrooms, where memory and connectivity determine which models can be run locally. Each model is assigned the smallest tier on which it could run as the primary foreground workload. Closed-weight models are placed in the cloud tier first, since their weights cannot be obtained at any size. The remainder are assigned on total parameter count against a per-tier memory budget, defined as the device's total RAM multiplied by the share a model can claim, at the precision assumed for that tier (4-bit on mobile and tablet, 8-bit at workstation class). The full methodology, assumptions and caveats are given in Appendix~\ref{app:tiers}.

\begin{figure}[t]
  \centering
  \includegraphics[width=0.85\linewidth]{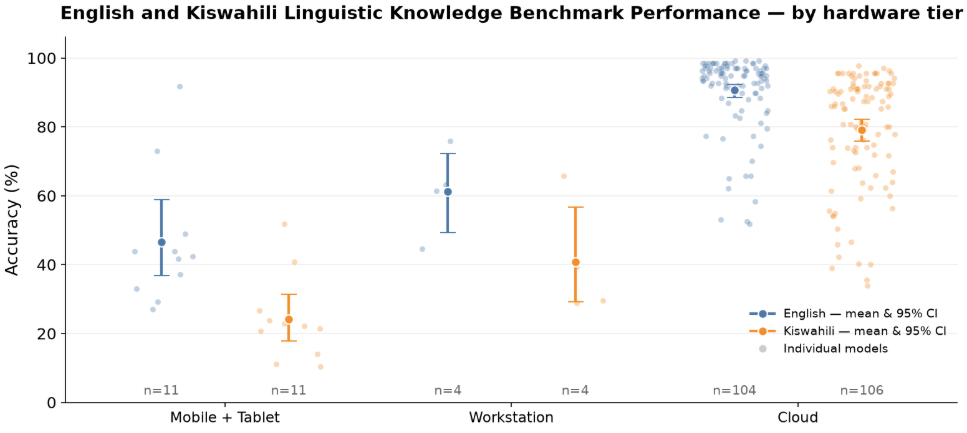}
  \caption{Benchmark accuracy by deployment hardware tier and language. Models are grouped by the smallest class of device on which they could run locally, estimated from total parameter count; mobile and tablet are reported together. Accuracy was measured at each model's best available precision (16-bit in almost all cases), whereas the tiers assume 4-bit quantisation on mobile and tablet and 8-bit at workstation class, so on-device accuracy would be below the values plotted here. Each marker is the tier's mean accuracy, with error bars giving a 95\% confidence interval from 1,000 bootstrap resamples over models; $n$ is the number of models in each tier.}
  \label{fig:tiers}
\end{figure}

Accuracy declines monotonically as the memory budget decreases, in both languages (Figure~\ref{fig:tiers}). Mean accuracy in the cloud tier is 91\% for English and 79\% for Kiswahili, falling to 61\% and 41\% at workstation class and to 47\% and 24\% on mobile and tablet hardware. The Kiswahili deficit persists at every tier and widens steadily as hardware shrinks, from 12 percentage points in the cloud tier to 20 at workstation class and 23 on mobile and tablet. The models most likely to be deployed in a low-resource classroom are thus the least capable in the language they would be deployed to serve. The workstation tier rests on 4 models and the mobile and tablet tier on 11, however, so their means should be read as indicative.

We did not run a formal human baseline, but our discussions with experts suggest that a trained Kiswahili lower-primary teacher would score roughly 80--90\% on these items. Relatedly, coverage across the nine linguistic domains and grade levels is intended to ensure the benchmark spans the curriculum, not to support performance claims within any individual domain or grade. Cell counts are small and unevenly distributed, so we do not interpret per-cell accuracies.

\section{Limitations}
\label{sec:limitations}

We identify three limitations of the present study, concerning the medium of the items, the scope of the construct, and the inference conditions.

\paragraph{Medium of the items.}
Questions are written in English and ask about the target language, which isolates linguistic knowledge from reading comprehension and instruction-following in the same language. Asking about a language in English is an easier task than operating in it, so this construction is likely to overstate what a model would achieve on the same content presented in the target language, and the near-ceiling scores we observe for frontier models may partly reflect the English medium. Results should be read as an upper bound. A version with items written in the target language would plausibly lower scores across the board, and the option of translating items programmatically remains open.

\paragraph{Scope of the construct.}
The benchmark is a retrieval test by design: it establishes whether a model holds a linguistic fact, not whether it can apply that fact in the situations a classroom tool would demand. This motivates its position as the first of three benchmarks in our suite: Linguistic Knowledge, Decodability Assessment, and Decodable Text Generation. The latter two test whether models can put that knowledge to work, by identifying words a learner cannot yet decode and by writing text within a given instructional sequence. A model can pass this benchmark while failing either. Retrieval is a necessary condition for decodable-text generation, not a sufficient one, and the near-ceiling frontier scores reported here should not be read as evidence that the end goal is met.

\paragraph{Inference conditions.}
All models are run under a single fixed configuration, and we did not explore the decoding and reasoning parameters that would plausibly change the ranking. Sampling settings such as temperature and other decoding controls were left at their defaults, which is likely to matter most for the smaller models, where answer stability is more sensitive to configuration. We inspected outputs to confirm that low scores were not artefacts of parsing or answer-extraction failure, so the gap we report is not a formatting artefact, but we cannot rule out that different decoding settings would raise smaller-model accuracy. Similarly, we did not vary reasoning effort or extended-thinking budgets for the larger models that support them, so their reported accuracy reflects one operating point rather than the frontier of what they can achieve. The reported figures should therefore be read as accuracy under one specific configuration, not as an upper bound on model capability.

\section{Conclusion}
\label{sec:conclusion}

We introduced the Linguistic Knowledge Benchmark, the first rung of a three-level evaluation ladder for decodable-text generation, and a reproducible pipeline that builds such a benchmark from a language's own curriculum and teaching materials with expert effort concentrated on the final review. Applied to Kiswahili and English, the benchmark separates models across nearly the full accuracy range. Frontier models answer almost all items correctly in both languages, but the cheaper and smaller models that could realistically run in low-resource classrooms lose much more ground in Kiswahili than in English, and this gap widens as hardware shrinks. For developers, the practical implication is that model choice for LMIC literacy tools cannot be made from English results alone. For the field, the results show where intervention is needed: smaller, deployable models need better coverage of the languages they would serve.

Because the benchmark is a retrieval test posed in English, it is a necessary but not a sufficient condition for producing decodable text. Next steps are versions with items written in the target language, extension of the pipeline to further LMIC languages, and the two upper rungs of the ladder, Decodability Assessment and Decodable Text Generation, which test whether models can apply this knowledge.

\section*{Data and Code Availability}

We open-source 20 items from each benchmark dataset on Hugging Face\footnote{\url{https://huggingface.co/collections/AI-for-Education/linguistic-knowledge-benchmark}} to make the item format and coverage inspectable, and hold the remainder back to limit leakage into training corpora. Model scores on both benchmarks are published on our public leaderboard.\footnote{\url{https://www.fab-ai.org/benchmarks?benchmarkType=foundational-literacy&benchmark=LK-sw&pageView=table&modelMode=default&dataset=&price=10&models=\%5B\%5D}} The full datasets are available to researchers on request from \href{mailto:maxime.lelievre@fab-ai.org}{\texttt{maxime.lelievre@fab-ai.org}}. All pipeline prompts are reproduced in Appendix~\ref{app:prompts}, and the code is open source on GitHub.\textsuperscript{\ref{fn:github}}

\section*{Acknowledgements}
We thank the pedagogy experts who reviewed the benchmark methodology and linguistic domains, and the native-speaking early-literacy specialists who reviewed every Kiswahili and English item. This includes: Anne Wanjiru Njogu, Marcia Davidson and Margaret (Peggy) Dubeck. Their judgement shaped the final datasets; any remaining errors are our own. We are also grateful to the curriculum developers, authors, teachers and publishers whose materials supported the generation of benchmark questions.

\bibliographystyle{plainnat}
\bibliography{references}

\clearpage
\appendix

\renewcommand{\thetable}{A\arabic{table}}
\renewcommand{\thefigure}{A\arabic{figure}}
\setcounter{table}{0}
\setcounter{figure}{0}

\section{Prompts}
\label{app:prompts}

The prompts reproduced below are the Kiswahili instantiation. The English instantiation shares the same structure (stages, domain spine, JSON schemas, rule numbering) and differs only in language-specific content: the target language and region, the grade vocabulary and reference curriculum, the domain glosses, and the worked examples and morphology-specific rules, which were rewritten for English as set out in Table~\ref{tab:adaptation}. Placeholders for attached files and run-time inputs are shown in angle brackets.

\begin{lstlisting}[caption={Pipeline Stage 1 (source survey).},label={prompt:stage1}]
You are a descriptive linguist and early-grade literacy specialist working on Swahili (Standard / Kenya). You are helping build a knowledge benchmark that tests whether a Swahili-capable language model holds the linguistic content knowledge a Primary literacy teacher needs. The benchmark is a RETRIEVAL test: each item probes one discrete, knowable linguistic fact -- not a teaching decision and not open-ended reasoning -- and a fluent Swahili speaker should be able to answer every item regardless of teaching experience.

Your job in THIS step is NOT to write questions. It is to SURVEY the attached document set -- going through each source one by one -- and summarise what each one contains that could be useful for building that benchmark, organised over a fixed grid of 9 linguistic domains x Primary grades G1-G5 (plus out-of-band G0 and beyond_primary). This map (the "digest") is the auditable evidence base a later step uses to decide how many questions to write where.

THE GRID (the context your survey is organised over)
DOMAINS (9; language-agnostic spine, instantiated for Swahili):
D1 Phonics & Orthography -- sound-letter correspondence, spelling rules, syllable structure
D2 Phonology & Suprasegmentals -- phoneme contrasts, length, stress/tone, prosody
D3 Nominal Morphology & Agreement -- number, gender/class, determiner/adjective/possessive agreement
D4 Verbal Morphology -- agreement, tense/aspect/mood, derivation & valency, non-finite forms, negation
D5 Syntax -- word order, argument structure, relativisation, focus, ellipsis, quantifiers
D6 Semantics & Lexicon -- word meaning, polysemy, semantic roles, scope
D7 Pragmatics & Discourse -- anaphora, reference tracking, information structure, narrative
D8 Sociolinguistic & Register -- honorifics, dialect, register, idiom/proverb
D9 Metalinguistic Reasoning -- rule induction from paradigms; explicit grammatical analysis

DOMAIN FILING (which domain a fact belongs to -- file by WHAT A SOLVER MUST KNOW to answer it, NOT by what the fact is superficially about):
- A Swahili grammatical / metalinguistic TERM (naming a rule, class, or structure) -> D9, regardless of the term's subject.
- What a word MEANS, or the word FOR something -> D6, regardless of the word's subject.
- Variation BETWEEN SPEAKERS (register, honorifics, proverbs, dialect, variety) -> D8. Boundary cases where the spine overlaps:
- Proverbs & fixed sayings -> D8. Figurative use of ordinary words -> D6.
- Discourse structures (how a text is built) -> D7. The Swahili LABELS for those structures -> D6. What a text is DOING -> D7; what a word/idiom MEANS -> D6 (D8 if a fixed saying).
- Register, honorifics, variety -> D8. Culturally-themed vocabulary -> D6. File a fact where the KNOWLEDGE lives; do not let its topic pull it into the wrong domain.

GRADES (anchor each fact to where it is FIRST introduced, even if the curriculum spirals back to it later in a harder form):
G0 = pre-primary / below Grade 1,
G1 = Grade 1   G2 = Grade 2   G3 = Grade 3   G4 = Grade 4   G5 = Grade 5
beyond_primary = above Grade 5
G1-G5 are the core Primary band; G0 and beyond_primary catch content that falls outside it. Record a fact at its true grade -- do NOT cram out-of-band content into G1/G5 or drop it.

DISCIPLINE (read first)
- Go through the sources ONE BY ONE: open and read EVERY attached source.
- Work ONLY from the attached sources. Your own prior knowledge of Swahili may be incomplete or subtly wrong and is NOT evidence.
- Every coverage claim carries a RESOLVABLE LOCATOR: source id + page/section + heading.
- "absent" / "not covered" is a REQUIRED, first-class value. Where a source is silent on a cell, say so. NEVER invent coverage to fill the grid.
- Count TESTABLE FACTS, not prose. A testable fact is one discrete, locatable fact you could write a single retrieval question from -- answerable by a fluent speaker.
- FILE EACH FACT by WHAT A SOLVER MUST KNOW to answer it, not by its topic (see DOMAIN FILING above). A term -> D9; a meaning/word -> D6; speaker-variation -> D8 -- even when the fact's subject matter belongs elsewhere.
- QUOTE A VERBATIM EXAMPLE for each testable fact, copied EXACTLY from this source (the target-language form/phrase + a short English gloss). COPY the form; never paraphrase it or reconstruct it from memory.
- ATTRIBUTE everything to a specific source. The point of the digest is to show what EACH source contributes, so a later run with a different set can be compared to it.
- When you place a fact at a grade, record HOW you know its grade using the grade_placement_evidence enum, honestly: language_empirical | related_language_extrapolated | curriculum_anchored | typologist_judgement | source_attested. If a grade is your own judgement rather than attested in a source, say "typologist_judgement" -- do NOT dress it up as "curriculum_anchored".

CLASSIFY EACH SOURCE BY TIER (the FIRST thing you do with a document, before its deep review)
For each document, before surveying its content, glance at its first few pages (title page, front matter, publisher, intended audience), identify WHAT KIND of document it is, record it as "tier", then go on to survey it: curriculum -- official syllabus, curriculum framework, scope-and-sequence (e.g. the national Kiswahili syllabus). teacher_material -- teacher's guides, lesson manuals, pacing charts. student_material -- pupil's books, graded readers, workbooks, exercise books. other -- anything else (e.g. a reference/descriptive grammar -- the "facts" source for verifying forms; dictionaries; academic papers). Give a one-line "tier_basis" citing what told you (e.g. "title page: 'Standard II Pupil's Book'").

Produce the resource digest as ONE strict JSON object:
"resource_digest":
{
  "by_document": [            // one entry per attached source, in the order you read them
    {
      "source_name":"<the file name of this source, used consistently in ALL locators>",
      "tier":"curriculum | teacher_material | student_material | other",
      "tier_basis":"<=12-word why, citing what you saw (title page, front matter, publisher, audience)>",
      "by_domain":[
        {
          "domain":"D1".."D9",
          "by_grade":[
            {
              "grade":"G0 | G1..G5 | beyond_primary",
              "coverage":"rich|some|thin|absent",
              "n_testable_facts_est":<int>,
              "example_facts":[       // one object per concrete testable fact
                {
                  "fact":"<=12-word phrasing of a concrete testable fact found here",
                  "example":"<verbatim target-language form/phrase quoted from THIS source> -- <short English gloss>"
                }
              ],
              "locators":["<source_name>:<page/section>:<heading>", ...],
              "grade_basis":"<grade_placement_evidence enum value> - <=8-word why>",
              "notes":"<caveats, or 'not covered'>"
            }
          ],
          "domain_summary":"<2-3 sentences: what THIS document offers in this domain - where it is rich vs thin, and at which grades its testable content concentrates>"
        }
      ],
      "document_summary":"<2-3 sentences: what this document as a whole brings - its strongest domains/grades and its main gaps>"
    }
  ]
}

RULES
- SURVEY ONLY. Do NOT write questions and do NOT assign item counts yet.
- Every `example_facts[].example` is a VERBATIM quote from this source (form + gloss), copied exactly -- not paraphrased or recalled.
- Testable-fact counts are estimates, but each must be backed by specific locators.
- grade_basis uses the grade_placement_evidence enum honestly (see DISCIPLINE).
- Cover the WHOLE grid: report thin and absent cells as explicitly as rich ones.
- Ground every claim in the attached documents and cite locations (source id + page/section + heading).
- Output ONLY the JSON object. No preamble, no markdown fences.

Produce the JSON object now.
\end{lstlisting}
\begin{lstlisting}[caption={Pipeline Stage 2 (feature map merge).},label={prompt:stage2}]
You are a descriptive linguist and benchmark designer working on Swahili (Standard / Kenya). You are helping build a knowledge benchmark that tests whether a Swahili-capable language model holds the linguistic content knowledge a Primary literacy teacher needs. The benchmark is a RETRIEVAL test: each item probes one discrete, knowable linguistic fact -- not a teaching decision and not open-ended reasoning -- and a fluent Swahili speaker should be able to answer every item regardless of teaching experience. Items sit on a fixed grid of 9 linguistic domains x Primary grades G1-G5 (plus out-of-band G0 and beyond_primary).

This is the MERGE step. A prior pass surveyed the sources and produced a "resource_digest": an auditable, PER-DOCUMENT map of what each attached source contains across that domain x grade grid, with per-grade testable facts, locators, and each document's tier. Your job is to CONSOLIDATE those per-document maps into ONE feature map -- grouped by domain, ordered by grade -- that the item-writing step builds questions from. You do NOT decide how many questions to write or which cells matter most; a later step does that. You only merge duplicate facts, organise, and keep every source visible.

THE GRID (fixed; the plan is organised over it)
DOMAINS (9; language-agnostic spine -- instantiate it for Swahili via the per-feature subdomain tags you add below, NOT by changing the spine): D1 Phonics/Orthography - D2 Phonology - D3 Nominal morphology - D4 Verbal morphology - D5 Syntax - D6 Semantics/Lexicon - D7 Pragmatics/Discourse - D8 Sociolinguistic/Register - D9 Metalinguistic reasoning.
GRADES (anchor each fact to where it is FIRST introduced; grades may spiral):
G0 = pre-primary / below Grade 1   G1 = Grade 1   G2 = Grade 2   G3 = Grade 3   G4 = Grade 4   G5 = Grade 5   beyond_primary = above Grade 5 (lower-secondary or scholarly)
grade_placement_evidence enum (how a grade claim is grounded -- record honestly): language_empirical | related_language_extrapolated | curriculum_anchored | typologist_judgement | source_attested

Produce ONE strict JSON object -- a "feature_map":
"feature_map":
{
  "by_domain":[
    {
      "domain":"D1".."D9",
      "features":[            // ALL features for this domain, ORDERED by grade: G0 -> G1..G5 -> beyond_primary
        {
          "subdomain":"<language-specific area WITHIN the domain, drawn from the sources, e.g. 'noun-class concord (cl. 7/8)' or 'applicative extension' -- free text, NOT a fixed list>",
          "grade":"G0 | G1..G5 | beyond_primary",
          "feature":"<short name>",
          "description":"<the rule/fact in English, precise enough to write a question from>",
          "example":"<verbatim Swahili example + English gloss>",
          "grade_placement_evidence":"<enum value>",
          "sources":[          // 1+ entries; MORE THAN ONE = the same fact attested by several documents
            {"locator":"<source_name>:<page/section>:<heading> -- must match a locator in the digest","tier":"curriculum | teacher_material | student_material | other"}
          ]
        }
      ]
    }
  ]
}

HOW TO MERGE (this is the whole job)
- INCLUDE EVERY testable fact the digest recorded. No counts, no importance ranking, no filtering -- that is a later step's job. A cell the digest marked "absent" simply yields no feature; never invent one to fill the grid.
- MERGE DUPLICATES -- ONE FEATURE PER FACT, NOT PER LABEL. If two or more documents attest the SAME underlying fact, emit ONE feature and list each document under "sources" (with its tier). This holds EVEN WHEN the sources name the fact differently or you would phrase it differently: merge on the fact itself, not on a matching label. Do NOT drop the extra source -- several sources backing a fact is a strength signal the item-writing step uses, so it must stay visible.
- CARRY THE TIER. Take each source's tier from the digest's by_document[].tier and put it on its "sources" entry. Every locator must match one already in the digest.
- GRADE ON CONFLICT. If the merged documents place the same fact at different grades, record the grade from the HIGHEST-TIER source (curriculum > teacher_material > student_material > other) and note the disagreement in the description.
- ORDER each domain's features by grade, easiest first (G0 -> G1..G5 -> beyond_primary).
- REFILE MISPLACED FACTS. File each feature by WHAT A SOLVER MUST KNOW to answer it, not by its topic: a Swahili grammatical/metalinguistic TERM -> D9; a word's MEANING or the word FOR something -> D6; variation BETWEEN SPEAKERS (register, honorifics, proverbs, variety) -> D8. Overlaps: proverbs/fixed sayings -> D8 but figurative use of ordinary words -> D6; discourse STRUCTURES -> D7 but their Swahili LABELS -> D6. If the digest placed a fact in a domain that contradicts this, MOVE it to the correct domain and note the move in the description. Grade is unchanged by a refile.
- SUBDOMAIN TAG. Tag each feature with an emergent subdomain so within-domain breadth is visible; the fixed D1-D9 spine does not change.
- CARRY THE EXAMPLE. Copy each feature's example verbatim from the digest fact it comes from (`example_facts[].example`). Do NOT re-quote from memory and do NOT invent one; the digest already captured it from the source.

RULES
- MERGE ONLY -- do NOT write questions, assign counts, or rank importance.
- Every feature traces to at least one digest locator, and every locator carries its tier.
- Honour the grade_placement_evidence enum honestly.
- Output ONLY the JSON object. No preamble, no markdown fences.

INPUTS
- RESOURCE DIGEST -- one block per document. It is the JSON file attached to this chat:
<<the resource_digest JSON file attached here>>

This is a JSON-to-JSON merge: build the feature_map from the attached digest alone.

Produce the JSON object now (feature_map).
\end{lstlisting}
\begin{lstlisting}[caption={Pipeline Stage 3 (item generation).},label={prompt:stage3}]
You are a linguist item-writer building a knowledge benchmark for Swahili (Standard / Kenya). The benchmark tests whether a Swahili-capable language model holds the linguistic content knowledge a Primary literacy teacher needs: how the language WORKS -- its phonics/orthography, phonology, morphology, syntax, lexicon and metalinguistic structure -- as opposed to its ability to USE the language. You write questions IN ENGLISH about Swahili.

THIS IS A SHARDED RUN. You are given the FULL feature_map but only a SUBSET of its cited sources (this shard). Generate candidate items ONLY for features that cite at least one of the ATTACHED sources; skip every other feature silently (it is handled in another shard). You are a CANDIDATE GENERATOR, not the final assembler: oversample and do not make any whole-dataset decisions (no target counts, no cross-domain balancing, no answer letter balancing, no final numbering) -- a later step selects and balances. Output a JSON array of candidate items for this shard.

CONTEXT (what kind of item to write)
This is a RETRIEVAL test: each item probes one discrete, knowable linguistic fact. A literate first-language speaker of the target language should be able to answer every item from knowledge of the language alone -- without pedagogical knowledge, world knowledge, or open-ended reasoning. Crucially, the item must require knowledge of THIS language: if it can be answered by general reasoning, by copying or segmenting a string shown in the stem, or by ranking the options without reading the stem, it is not a knowledge item.

THE GRID (assigned upstream; carry both onto each item)
DOMAINS (language-agnostic spine, instantiated for Swahili):
D1 Phonics & Orthography  D2 Phonology & Suprasegmentals  D3 Nominal Morphology & Agreement  D4 Verbal Morphology  D5 Syntax  D6 Semantics & Lexicon  D7 Pragmatics & Discourse  D8 Sociolinguistic & Register  D9 Metalinguistic Reasoning
GRADES -- where the content is FIRST introduced (even if revisited later in harder form): G1-G5 = Primary Grades 1-5; G0 = pre-primary; beyond_primary = above Grade 5. Grade marks the content level, not the item's difficulty (see DIFFICULTY).

FRAMEWORK TAG (literacy-construct layer; tag EVERY item)
Each item already carries a linguistic domain (D1-D9). This tag records, separately, which strand of the Simple View of Reading (Gough & Tunmer, 1986) the tested knowledge serves, and which GEEAP (2025) core instructional skill an error in it would corrupt. The layers are independent: the domain says what KIND of fact it is; the framework tag says what the fact is FOR in learning to read.

THE FUNCTIONAL TEST (apply this, not the domain label)
Ask: in a child's reading, does this knowledge serve turning WRITTEN FORMS into words (decoding), or turning words into MEANING (language_comprehension)?
- "decoding": letter-sound mapping, spelling conventions, syllable structure, sound contrasts as written, segmenting written word forms into readable parts (including morpheme-level segmentation of long agglutinative words).
- "language_comprehension": what forms MEAN or how they AGREE -- concord, tense/aspect semantics, word meaning, sentence interpretation, register, discourse. Oral-language content (greetings, exchange patterns, proverbs the curriculum teaches) belongs HERE, under geeap_component oral_language -- early-grade curricula carry it inside the literacy programme.
- "outside_flr_construct": the knowledge has NO bearing on supporting a child's literacy (e.g. language-family classification, dialectology, etymology of linguistic labels). If a fact only supports an outside_flr_construct item, DO NOT WRITE it -- skip the feature. This tag should therefore be rare-to-absent in your output; use it only to justify skipping, never to ship an item.

PER-DOMAIN CROSSWALK (defaults + overrides for svr_strand and geeap_component)
The two axes are INDEPENDENT: svr_strand records what the knowledge is FOR in learning to read; geeap_component records the lesson TYPE an error would corrupt. Start from the domain's default row; apply an override ONLY when its stated condition is met. The FUNCTIONAL TEST above is the tie-breaker, never the domain label alone.
D1 Phonics & Orthography -- default svr=decoding, geeap=systematic_phonics. override: spelling/punctuation conventions (apostrophe in ng', number-words, Arabic-origin spellings) -> geeap=writing. Segmentation/blending stay systematic_phonics.
D2 Phonology & Suprasegmentals -- default svr=decoding, geeap=phonological_awareness. override: stress/prosody items -> geeap=reading_fluency (svr STAYS decoding).
D3 Nominal Morphology -- default svr=language_comprehension, geeap=oral_language. override: vowel-merger morphophonology (written form) -> svr=decoding, geeap=systematic_phonics. An irregular plural is NOT a merger.
D4 Verbal Morphology -- default svr=language_comprehension, geeap=oral_language. override: verb-morpheme SEGMENTATION -> svr=decoding, geeap=systematic_phonics. Building/assembling a form is NOT segmentation (it stays the default).
D5 Syntax -- default svr=language_comprehension, geeap=oral_language. override: interpreting PASSAGE-level structure -> geeap=reading_comprehension. Sentence-level grammar does NOT qualify.
D6 Semantics & Lexicon -- default svr=language_comprehension, geeap=oral_language. override: polysemy/figurative meaning that requires RESOLVING the sense in context -> geeap=reading_comprehension. Simply reporting that a word is polysemous stays oral_language.
D7 Pragmatics & Discourse -- default svr=language_comprehension, geeap=reading_comprehension. override: direct vs reported speech with a PUNCTUATION focus -> geeap=writing.
D8 Sociolinguistic & Register -- default svr=language_comprehension, geeap=oral_language. override: proverbs met IN A READING PASSAGE the item asks the reader to interpret -> geeap=reading_comprehension. Dialectology / language classification -> svr=outside_flr_construct, geeap=none.
D9 Metalinguistic Reasoning -- svr and geeap INHERIT the rule the item is about: a spelling/letter term -> decoding/systematic_phonics; a punctuation term -> decoding/writing; a fluency/prosody term -> decoding/reading_fluency; a grammar/agreement/word-class term -> language_comprehension/oral_language.

APPLYING AN OVERRIDE
- An override fires ONLY when its stated condition is met. If not met, use the default. Do NOT apply an override just because a keyword matches.
- geeap "none" is valid ONLY together with svr=outside_flr_construct.

COVERAGE (this shard)
A feature is ELIGIBLE only if at least one source in its `sources` field is a file ATTACHED to this prompt (match `sources[].locator`'s source_name against the attached filenames). Generate candidates for every eligible feature and for NO other -- if none of a feature's sources are attached, skip it silently (another shard handles it). For each eligible feature write 1-K candidates (default K=2): one when the feature supports a single good item, up to K when it spans genuinely distinct sub-cases (e.g. the same concord in two different noun classes) -- each candidate uses a different example and frame, not a reworded twin. Choose answer_type per candidate to suit the fact (production facts suit short_form; contrasts and judgements suit MCQ). Do NOT assign per-domain counts or balance grades, answer types, or answer letters across the set.

DIFFICULTY REFLECTS TEACHER KNOWLEDGE, NOT STUDENT LEVEL. The grade sets the content; the item must be pitched at the teacher responsible for that grade and assess command of that content -- applying a rule to a real form, an irregular formation, or a distinction commonly mistaken -- not student-level recall or anything recoverable from the stem.

FEATURE FIDELITY. Copy `linguistic_feature` VERBATIM and test exactly the fact it names; if an item drifts to a different fact, rewrite it back or drop it. Check the feature is filed by WHAT MUST BE KNOWN (a term -> D9, a meaning/word -> D6, speaker-variation -> D8); if its domain looks wrong, skip it.

OUTPUT -- a JSON array of candidate item objects (no global numbering yet):
{
  "shard_id": "<short label for this shard, e.g. the attached-doc set name>",
  "candidate_id": "<unique within this shard, e.g. c001, c002 ...>",
  "domain": "<carried from the feature>",
  "subdomain": "<carried from the feature>",
  "grade": "<carried; G0 | G1..G5 | beyond_primary>",
  "answer_type": "multiple_choice" | "short_form",
  "question": "<English question; embed target-language material verbatim>",
  "options": {"A":"...","B":"...","C":"...","D":"..."},   // MCQ only; omit for short_form
  "expected_answers": ["B"],             // MCQ: one letter. short_form: list of accepted strings
  "linguistic_feature": "<carried from the feature>",
  "svr_strand": "decoding | language_comprehension | outside_flr_construct",
  "geeap_component": "systematic_phonics | phonological_awareness | oral_language | reading_fluency | reading_comprehension | writing | none",
  "sources": [ {"locator":"...","tier":"..."} ],   // carried; the one(s) you verified against
  "grade_placement_evidence": "<carried from the feature>"
}

ANTI-GAMING RULES (every item must pass ALL of these)
1. NO ANSWER IN THE STEM (the most important rule). The word, morpheme, or letters the item asks for must NOT be visible in any form shown in the stem -- including inside /phoneme slashes/, in a gloss, or as separable parts the solver only has to join. If the answer can be read off, copied, or segmented out of what the stem prints, the item is broken.
   - Wrong: "In atakaa, which marker shows the future?" (-ta- is sitting in atakaa).
   - Right: give the parts and require the whole -- "Mark the verb -soma for third-person-singular future; write the full form." (answer atasoma, not shown).
2. DON'T STATE OR DEMONSTRATE THE TESTED RULE. Do not name the class or rule and then ask for the very form it dictates, and do not show a worked example of the same transformation the item asks for. Covers: stating "M-WA" then asking for the wa- form; the alliterative-concord trap (matching the adjective prefix to a noun prefix already on screen); and rule-induction items that hand a complete regular paradigm.
3. USE THE CASE WHERE SURFACE-MATCHING FAILS. For morphology and concord, prefer an irregular, an exception, or a NON-alliterative class (where the agreement prefix differs from the visible noun prefix), so a reader who only "makes it rhyme" or extrapolates the pattern gets it wrong. Regular, transparent, alliterative examples do not test knowledge.
4. LENGTH PARITY (hard constraint). All four options within ~15 characters of one another; the correct option must NOT be the longest or the most complete. No "kitchen-sink" option that bundles two or more ideas unless all four do. Match hedging and jargon across all four (no "typically/often" or stacked technical terms only on the key).
5. UNIFORM OR ABSENT GLOSSES. A parenthetical gloss or label must appear on ALL four options or on none. An option's gloss must never name the property being tested (do not tag only the correct option "(animate)" when the stem is about animacy).
6. PARALLEL OPTIONS. All four options the same grammatical form and register; if they are competing forms, keep them minimal so the linguistic contrast is the only signal. No option a substring of another. No absolutist words ("always/never/only") in distractors only.
7. DISTRACTORS MUST FUNCTION. Each distractor encodes a distinct, nameable error a real learner or speaker would make (the wrong concord, the wrong TAM marker, an over-regularised plural). No distractor should be eliminable just by reading the stem, and none should be implausibly absurd. Exactly one correct option; no true/false, no "all/none of the above".
8. SHORTCUT CHECK (apply before finalising each MCQ). Cover the stem and look only at the options: if the answer is still identifiable -- longest, only hedged/glossed/complete one, or the only one echoing the topic -- rewrite it.
9. CONSTRUCT CHECK (sharpened). The item must turn on a fact about THIS language. Reject and rewrite anything answerable by: concatenating syllables, reading letters off a shown word, general literacy, or general grammar/linguistics not specific to the target language.
10. STEM STANDS ALONE. The stem must be self-contained: never reference a document, curriculum, grade, or lesson (no "as in the syllabus", "at this grade", "in the reading").
11. NO SCAFFOLDING CLUE. A descriptive hint in the stem must not make the answer findable without the tested knowledge. Wrong: asking for a word while stating it "has a doubled vowel in each of its two parts" -- that describes the answer's shape and lets a solver reverse-engineer it. Describe the target by its function, not by features that reveal it.

POSITIVE TEMPLATES (use these forms, especially in the form domains D1-D4)
- PRODUCTION: give the morphemes/parts, require the assembled form (encoding, not recognition). Short_form suits this.
- EXCEPTION: target an irregular or a form where the regular pattern over-applies.
- JUDGEMENT: ask which form is correct or which is wrong (grammaticality, spelling), so the answer cannot be lifted off a string.
- CONTRAST: a minimal pair where the surface does not disambiguate ("which begins with a single nasal rather than nasal-plus-g, ng'ombe or ngao?").

CARRY-OVER RULES
- ENGLISH MEDIUM; only quoted target-language forms appear in Swahili. Plain language, no rare metalinguistic jargon unless it is the thing being tested.
- TEXT-ONLY, IMPERSONAL: phrase every stem impersonally (no "you"); never require the model to speak, listen, or hear. Test only what is recoverable from the written form.
- TRANSLATION ANSWER-NEUTRAL: a short English gloss may aid clarity but must not reveal the tested answer; for D6 meaning items, omit the gloss of the queried word.
- GROUNDING: every item supported by one of its feature's sources; carry that (locator, tier) onto the item and verify forms against the ATTACHED source, not memory. On disagreement, ground in the highest tier (curriculum > teacher_material > student_material > other).
- LETTER BALANCE: spread the correct MCQ answer roughly evenly across A-D over the set.
- SHORT_FORM: answer 1-3 words; list all accepted spelling/spacing/hyphen variants in expected_answers (not synonyms or paraphrases).
- Output ONLY the JSON array. No preamble, no markdown fences.

INPUTS
- SOURCES: this shard's subset of the cited PDFs. Only features citing one of THESE are eligible.
- FEATURE_MAP -- the full feature_map (all 9 domains; features grade-ordered). It is the JSON file attached to this chat. Carry subdomain / grade / sources / grade_placement_evidence onto each candidate, and assign svr_strand and geeap_component using the FRAMEWORK TAG block above:
<<the feature_map JSON file attached here>>

Produce the JSON array of candidate items for THIS shard now.
\end{lstlisting}
\begin{lstlisting}[caption={Pipeline Stage 4 (quality control and deduplication).},label={prompt:stage4}]
You are a benchmark editor reviewing candidate items for a Swahili (Standard / Kenya) linguistic-knowledge benchmark. Several sharded generation runs each produced candidate items from different source subsets. POOL them, QUALITY-CHECK each one, and mark DUPLICATES -- without rewriting, adding, or deleting any candidate. Every input candidate appears exactly once in your output, annotated with three new fields.

You work from the candidate TEXT alone -- no sources are attached. Judge only what is checkable from the item itself (the shape / anti-gaming rules below), NOT whether a Swahili form is faithful to a source (that was verified at generation).

QUALITY CHECK -- score each candidate against these rules:
1. NO ANSWER IN THE STEM -- the asked-for form/morpheme/letters are not visible in the stem in ANY form, including inside /phoneme slashes/, a gloss, or as separable parts to join.
2. RULE NOT STATED OR DEMONSTRATED -- the item does not name the rule/class then ask for the form it dictates, and shows no worked twin of the transformation it asks for.
3. LENGTH PARITY (MCQ) -- the four options are within ~15 characters; the key is not the longest or most complete; hedging/jargon matched across all four.
4. UNIFORM OR ABSENT GLOSSES (MCQ) -- a gloss/label is on all four options or none, and never names the tested property.
5. PARALLEL OPTIONS (MCQ) -- all four the same form/register; none a substring of another; no absolutist words ("always/never/only") in distractors only.
6. DISTRACTORS FUNCTION (MCQ) -- each distractor is a distinct, nameable error; none absurd; exactly one correct; no true/false, no "all/none of the above".
7. SHORTCUT CHECK (MCQ) -- with the stem covered, the key is not identifiable by length, gloss, completeness, or topic echo.
8. CONSTRUCT CHECK -- the item turns on a fact about Swahili, not general reasoning, string-copying, or general literacy/linguistics.
9. WELL-FORMED -- required fields present for its answer_type (MCQ: 4 options + exactly one letter in expected_answers; short_form: no options, 1-3 word expected_answers).
10. ENGLISH MEDIUM -- the question is written in English; only quoted target-language forms appear in Swahili, and no rare metalinguistic jargon is used unless it is the very thing being tested.
11. TEXT-ONLY & IMPERSONAL -- the stem is phrased impersonally (no "you") and never asks the reader to speak, listen, or hear; everything tested is recoverable from the written form.
12. TRANSLATION ANSWER-NEUTRAL -- any English gloss aids clarity without revealing the tested answer; for a meaning item, the queried word is not glossed.
13. STEM STANDS ALONE -- the stem references no document, curriculum, grade, or lesson.
14. NO SCAFFOLDING CLUE -- no descriptive hint in the stem makes the answer findable without the tested knowledge (e.g. stating the target word's letter/shape pattern).
15. IN-CONSTRUCT -- svr_strand is not "outside_flr_construct" (such items should not exist; fail any that do).
Rules 3-7 apply to multiple_choice items only; short_form items are exempt from them. Rules 1-2 and 8-15 apply to every item.

DEDUPLICATE -- ONE GROUP PER FACT, NOT PER FEATURE LABEL:
- Give every candidate a "dup_group_id". Candidates that probe the SAME underlying fact share one id -- judge this on the fact the item actually tests (its stem + answer), NOT on a matching `linguistic_feature` label. Two candidates with DIFFERENT feature labels still share an id if they test the same fact; two with the SAME label are separate groups if they test genuinely different facts. A candidate with no twin is its own group of one.
- Within each group set "quality_rank" = 1 for the best candidate (most rules passed, best-formed, hardest to game), 2 for the next, and so on. Rank on quality only; do NOT drop the losers -- the assembler keeps them as backups.

Produce ONE strict JSON array -- every input candidate, unchanged except for three ADDED fields:
{
  ...all fields of the input candidate, VERBATIM (shard_id, candidate_id, domain, subdomain, grade, answer_type, question, options, expected_answers, linguistic_feature, svr_strand, geeap_component, sources, grade_placement_evidence)...,
  "qc_verdict": {
    "pass": true|false,            // false if ANY applicable rule fails
    "failed_rules": [<numbers of the rules that failed>],
    "note": "<=15-word reason if failed, else ''>"
  },
  "dup_group_id": "<id shared by same-fact candidates, e.g. g001>",
  "quality_rank": <int within the dup group; 1 = best>
}

RULES
- ANNOTATE ONLY -- never edit a candidate's text, add candidates, or remove any. Count in == count out.
- Output ONLY the JSON array. No preamble, no markdown fences.

INPUTS
- CANDIDATES -- the candidate items from the generation runs. They are the JSON file(s) attached to this chat (one file per generation run, or a single combined file; several files may be attached):
<<the candidate JSON file(s) attached here>>

Produce the annotated JSON array now.
\end{lstlisting}
\begin{lstlisting}[caption={Question-answering prompts used at evaluation for multiple-choice items (top) and short-answer items (bottom). LANGUAGE is substituted with the target language at run time (e.g.\ Swahili, English).},label={prompt:answer}]
=== Evaluation -- Question Answering (multiple choice) ===
The following question is written in English and concerns the linguistics of LANGUAGE.
Answer the multiple-choice question by selecting the letter corresponding to the correct answer (A, B, C or D).
Only provide the letter for your answer.
Stop exactly after the letter.

<question and options (to answer)>

=== Evaluation -- Question Answering (short form) ===
The following question is written in English and concerns the linguistics of LANGUAGE.
Answer the question.
Provide only the answer in 1-3 words. Do not explain.

<question (to answer)>
\end{lstlisting}
\begin{lstlisting}[caption={LLM-judge prompts used at evaluation for short-answer items (top) and for multiple-choice items whose response could not be parsed by regular expression (bottom). Each returns a boolean verdict with a short justification, enforced with structured outputs.},label={prompt:judge}]
=== Evaluation -- LLM Judge (short form) ===
You are an expert evaluator for LANGUAGE language knowledge questions. You will be given a question, the model's answer, and one or more acceptable expected answers.

Your task is to determine if the model's answer matches one of the expected answers. Be strict: only accept answers that match one of the provided expected answers. Do NOT accept synonyms, paraphrases, or alternative forms that are not listed.

The only acceptable variations are: capitalization, leading/trailing hyphens and spacing.

Question: <the item's question, as asked>
Model's answer: <the model's raw output>
Acceptable expected answers: <expected answers, comma-separated>
Answer type: short_form
Is the model's answer correct?

=== Evaluation -- LLM Judge (multiple-choice fallback) ===
You are an expert evaluator. A model was asked a question and its response could not be parsed by a simple regex.

You will be given the question, the model's raw response, the expected answer, and the answer type (multiple_choice).

For multiple-choice: determine if the model's response indicates the correct letter (A, B, C, or D), even if the response contains extra explanation or formatting.

Respond with whether the model's answer is correct.

Question: <the item's question, as asked>
Model's answer: <the model's raw output>
Acceptable expected answers: <expected answers, comma-separated>
Answer type: multiple_choice
Is the model's answer correct?
\end{lstlisting}

\clearpage
\section{Supplementary Tables}
\label{app:tables}

\begin{table}[h]
  \centering
  \small
  \begin{minipage}[t]{0.40\linewidth}
    \centering
    \textbf{(a) Linguistic domains}\\[4pt]
    \begin{tabular}{ll}
      \toprule
      Domain & Covers \\
      \midrule
      D1 & Phonics and orthography \\
      D2 & Phonology \\
      D3 & Nominal morphology \\
      D4 & Verbal morphology \\
      D5 & Syntax \\
      D6 & Semantics and lexicon \\
      D7 & Pragmatics and discourse \\
      D8 & Sociolinguistics and register \\
      D9 & Metalinguistic reasoning \\
      \bottomrule
    \end{tabular}
  \end{minipage}\hfill
  \begin{minipage}[t]{0.58\linewidth}
    \centering
    \textbf{(b) Grade levels}\\[4pt]
    \setlength{\tabcolsep}{4pt}
    \begin{tabular}{l>{\raggedright\arraybackslash}p{2.0cm}>{\raggedright\arraybackslash}p{2.6cm}}
      \toprule
      Grade tag & Kiswahili (Kenya) & English (England) \\
      \midrule
      \texttt{G0} & Pre-primary & Reception / below Year 1 \\
      \texttt{G1} & Grade 1 & Year 1 \\
      \texttt{G2} & Grade 2 & Year 2 \\
      \texttt{G3} & Grade 3 & Year 3 \\
      \texttt{G4} & Grade 4 & Year 4 \\
      \texttt{G5} & Grade 5 & Year 5 \\
      \texttt{beyond\_primary} & Above Grade 5 & Above Year 5 \\
      \bottomrule
    \end{tabular}
  \end{minipage}
  \caption{The fixed grid on which every benchmark item sits. (a) The nine linguistic domains. (b) Grade tags. The grid is held constant across languages so that item sets remain comparable.}
  \label{tab:grid}
\end{table}

\begin{table}[h]
  \centering
  \footnotesize
  \begin{tabularx}{\linewidth}{>{\raggedright\arraybackslash}p{3.4cm}>{\raggedright\arraybackslash}X>{\raggedright\arraybackslash}X}
    \toprule
     & Kiswahili & English \\
    \midrule
    \multicolumn{3}{l}{\textit{Language-specific content (adapted)}} \\
    Language and region & Swahili (Standard / Kenya) & English (United Kingdom) \\
    Grade vocabulary & Grades 1--5; pre-primary; above Grade 5 & Years 1--5; Reception; above Year 5 \\
    Reference curriculum & National Kiswahili syllabus & National Curriculum for England \\
    Domain glosses (D2, D3, D4, D8) & stress/tone; gender/class; tense/aspect/mood, valency; honorifics & stress/intonation; plural formation; tense/aspect, derivation; formality \\
    Worked examples in rules & \textit{atakaa} / \textit{atasoma}; \textit{ng'ombe} vs \textit{ngao}; ``animate'' & \textit{stopped} / \textit{stop}; \textit{comb} vs \textit{card}; ``uncountable'' \\
    Morphology-specific rules & noun-class and alliterative concord; agglutinative segmentation; vowel-merger morphophonology & spelling rules and irregular plurals; multi-morpheme segmentation (\textit{un-lock-ing}); consonant doubling, \textit{y}$\to$\textit{i} \\
    Glossing convention & verbatim form + English gloss; ``translation answer-neutral'' & verbatim word/phrase; ``answer-neutral glossing'' \\
    \midrule
    \multicolumn{3}{l}{\textit{Pipeline scaffolding (unchanged)}} \\
    Stages & Four prompted stages; IRT-inspired selection by script & \textit{identical} \\
    Domain spine & D1--D9, language-agnostic & \textit{identical} \\
    Grade grid & G0, G1--G5, beyond\_primary & \textit{identical} \\
    Output schemas & \texttt{resource\_digest}, \texttt{feature\_map}, candidate item, QC verdict & \textit{identical} \\
    Rule sets & 11 anti-gaming rules; 15 quality-control rules & \textit{identical} \\
    \bottomrule
  \end{tabularx}
  \caption{Adaptation of the pipeline prompts from Kiswahili to English.}
  \label{tab:adaptation}
\end{table}

\begin{table}[h]
  \centering
  \small
  \begin{tabular}{llr}
    \toprule
    Model & Provider & Size (B params) \\
    \midrule
    Claude Haiku 4.5 & Anthropic & --- \\
    Claude Sonnet 4.6 & Anthropic & --- \\
    Gemini-3.1 Flash-Lite & Google & --- \\
    Gemini-3.1 Pro & Google & --- \\
    Gemma-3 12B & Google & 12 \\
    Gemma-4 31B & Google & 31 \\
    Ministral 3 8B & Mistral & 8 \\
    Mistral Small 4 & Mistral & 119 \\
    GPT-4.1 & OpenAI & --- \\
    GPT-5.4 Nano & OpenAI & --- \\
    \bottomrule
  \end{tabular}
  \caption{Model panel used for item selection. The 10 models span four providers and include both open-weight and proprietary models. Parameter counts are given only for open-weight models.}
  \label{tab:panel}
\end{table}

\begin{table}[h]
  \centering
  \small
  \begin{tabular}{lrr}
    \toprule
     & English & Kiswahili \\
    \midrule
    Panel (models / providers) & 10 / 4 & 10 / 4 \\
    Candidate items & 377 & 307 \\
    \quad of which scored by the panel & 370 & 298 \\
    Excluded (all models correct) & 217 & 85 \\
    Selected for review & 140 & 150 \\
    \midrule
    \textbf{Overall retention (of candidates)} & \textbf{37.1\%} & \textbf{48.9\%} \\
    \textbf{Retention range, domains} & \textbf{16.0--71.8\%} & \textbf{20.0--59.0\%} \\
    \textbf{Retention range, grades} & \textbf{24.3--54.1\%} & \textbf{0--58.2\%} \\
    \midrule
    Spearman $\rho$ / Kendall $\tau$ & 1.00 / 1.00 & 1.00 / 1.00 \\
    Mean leave-two-out selection frequency & 0.90 & 0.89 \\
    \bottomrule
  \end{tabular}
  \caption{Item selection and validation summary for English and Kiswahili. Item selection requires a complete response matrix, so 7 English and 9 Kiswahili items were excluded where any model's multiple-choice response could not be parsed.}
  \label{tab:selection}
\end{table}

\begin{table}[h]
  \centering
  \scriptsize
  \begin{tabularx}{\linewidth}{>{\ttfamily}p{3.7cm}>{\raggedright\arraybackslash}p{3.9cm}>{\raggedright\arraybackslash}X}
    \toprule
    \normalfont Field & Type / values & Description \\
    \midrule
    question\_id & int; 1--N, contiguous, unique & Question identifier \\
    domain & D1--D9 & Linguistic domain the item is filed under \\
    subdomain & free text & Language-specific sub-topic within the domain \\
    grade & G0--G5, beyond\_primary & Grade at which the content is first introduced \\
    answer\_type & multiple\_choice, short\_form & Item format \\
    question & text & Question stem \\
    options & dict; keys A--D & Answer options; present only when \texttt{answer\_type} = \texttt{multiple\_choice} \\
    expected\_answers & list[str] & Accepted answers: exactly one letter for multiple\_choice; 1--3 strings for short\_form \\
    linguistic\_feature & text & The specific linguistic fact the item tests \\
    svr\_strand & decoding, language\_comprehension, outside\_flr\_construct & Simple View of Reading strand the tested knowledge serves \\
    geeap\_component & systematic\_phonics, phonological\_awareness, oral\_language, reading\_fluency, reading\_comprehension, writing, none & GEEAP core instructional skill an error in the item would corrupt \\
    sources & list of \{locator, tier\} & Source material the item is grounded in (1--3 per item); tier is one of curriculum, teacher\_material, student\_material, other \\
    grade\_placement\_evidence & curriculum\_anchored, language\_empirical, source\_attested, typologist\_judgement, related\_language\_extrapolated & How the grade assignment was justified in the pipeline \\
    \bottomrule
  \end{tabularx}
  \caption{Structure of a benchmark item. The same schema applies to both languages.}
  \label{tab:schema}
\end{table}

\begin{table}[!h]
  \centering
  \footnotesize
  \setlength{\tabcolsep}{4pt}
  \urlstyle{same}
  \begin{tabularx}{\linewidth}{@{}r>{\raggedright\arraybackslash}p{0.40\linewidth}>{\raggedright\arraybackslash}X@{}}
    \toprule
    \# & Title & Link \\
    \midrule
    1 & Kiswahili 1: Kitabu cha Mwanafunzi & \href{https://shared.rti.org/content/tusome-grade-1-pupils-book}{\nolinkurl{shared.rti.org/content/tusome-grade-1-pupils-book}} \\
    2 & Kiswahili 1: Mwongozo wa Mwalimu & \href{https://shared.rti.org/content/tusome-grade-1-teachers-guide}{\nolinkurl{shared.rti.org/content/tusome-grade-1-teachers-guide}} \\
    3 & Kiswahili 2: Kitabu cha Mwanafunzi & \href{https://shared.rti.org/content/tusome-class-2-pupil-book-tusome-early-literacy-programme}{\nolinkurl{shared.rti.org/content/tusome-class-2-pupil-book-tusome-early-literacy-programme}} \\
    4 & Kiswahili 2: Mwongozo wa Mwalimu & \href{https://www.maktaba.org/book/1695/kiswahili-mazoezi-ya-lugha-mwongozo-wa-mwalimu-darasa-la-2}{\nolinkurl{www.maktaba.org/book/1695/kiswahili-mazoezi-ya-lugha-mwongozo-wa-mwalimu-darasa-la-2}} \\
    5 & Kiswahili 3: Mazoezi ya Lugha, Kitabu cha Mwanafunzi, Darasa la 3 & \href{https://www.maktaba.org/book/1693/kiswahili-mazoezi-ya-lugha-kitabu-cha-mwanafunzi-darasa-la-3}{\nolinkurl{www.maktaba.org/book/1693/kiswahili-mazoezi-ya-lugha-kitabu-cha-mwanafunzi-darasa-la-3}} \\
    6 & Kiswahili 3: Mazoezi ya Lugha, Mwongozo wa Mwalimu, Darasa la 3 & \href{https://www.maktaba.org/book/1696/kiswahili-mazoezi-ya-lugha-mwongozo-wa-mwalimu-darasa-la-3}{\nolinkurl{www.maktaba.org/book/1696/kiswahili-mazoezi-ya-lugha-mwongozo-wa-mwalimu-darasa-la-3}} \\
    7 & Lower Primary Level Curriculum Designs, Volume One (KICD, 2017) & \href{https://kicd.ac.ke/wp-content/uploads/2017/10/Volume-1-curriculum-designs-September-2017.pdf}{\nolinkurl{kicd.ac.ke/wp-content/uploads/2017/10/Volume-1-curriculum-designs-September-2017.pdf}} \\
    8 & Grade 4 Curriculum Designs, Volume 1 (KICD, 2019) & \href{https://github.com/jamilo-school/landing-page/blob/main/CurriculumDesigns/Grade\%204/volume-one.pdf}{\nolinkurl{github.com/jamilo-school/landing-page/blob/main/CurriculumDesigns/Grade\%204/volume-one.pdf}} \\
    9 & Grade 5 Curriculum Design: Kiswahili (KICD, 2019) & \href{https://github.com/jamilo-school/landing-page/blob/main/CurriculumDesigns/Grade\%205/KISWAHILI.pdf}{\nolinkurl{github.com/jamilo-school/landing-page/blob/main/CurriculumDesigns/Grade\%205/KISWAHILI.pdf}} \\
    10 & Swahili Learners' Reference Grammar (Thompson \& Schleicher, 2001) & \href{https://www.maktaba.org/book/1236/swahili-learners-reference-grammar}{\nolinkurl{www.maktaba.org/book/1236/swahili-learners-reference-grammar}} \\
    11 & Swahili Grammar for Introductory and Intermediate Levels (Almasi, Fallon \& Wared, 2014) & \href{https://www.bloomsbury.com/us/swahili-grammar-for-introductory-and-intermediate-levels-9780761879817/}{\nolinkurl{www.bloomsbury.com/us/swahili-grammar-for-introductory-and-intermediate-levels-9780761879817/}} \\
    \bottomrule
  \end{tabularx}
  \caption{Source documents used to build the Kiswahili benchmark (Standard Kiswahili, Kenya).}
  \label{tab:sources-sw}
\end{table}

\begin{table}[!h]
  \centering
  \footnotesize
  \setlength{\tabcolsep}{4pt}
  \urlstyle{same}
  \begin{tabularx}{\linewidth}{@{}r>{\raggedright\arraybackslash}p{0.40\linewidth}>{\raggedright\arraybackslash}X@{}}
    \toprule
    \# & Title & Link \\
    \midrule
    1 & English programmes of study: key stages 1 and 2 (National curriculum, September 2013) & \href{https://assets.publishing.service.gov.uk/media/5a7de93840f0b62305b7f8ee/PRIMARY_national_curriculum_-_English_220714.pdf}{\nolinkurl{assets.publishing.service.gov.uk/media/5a7de93840f0b62305b7f8ee/PRIMARY_national_curriculum_-_English_220714.pdf}} \\
    2 & 2026 KS1 English grammar, punctuation and spelling test mark schemes & \href{https://assets.publishing.service.gov.uk/media/6a156cca0026f30a6d421cf3/2026_KS1_English_GPS_test_mark_schemes.pdf}{\nolinkurl{assets.publishing.service.gov.uk/media/6a156cca0026f30a6d421cf3/2026_KS1_English_GPS_test_mark_schemes.pdf}} \\
    3 & 2026 KS2 English grammar, punctuation and spelling mark schemes & \href{https://assets.publishing.service.gov.uk/media/6a0af15cc510c3913d826736/2026_KS2_English_GPS_mark_schemes.pdf}{\nolinkurl{assets.publishing.service.gov.uk/media/6a0af15cc510c3913d826736/2026_KS2_English_GPS_mark_schemes.pdf}} \\
    4 & 2026 phonics screening check: scoring guidance & \href{https://assets.publishing.service.gov.uk/media/6a327de2e590e5e061c9a130/2026_phonics_screening_check_scoring_guidance.pdf}{\nolinkurl{assets.publishing.service.gov.uk/media/6a327de2e590e5e061c9a130/2026_phonics_screening_check_scoring_guidance.pdf}} \\
    5 & Letters and Sounds: Principles and Practice of High Quality Phonics (DfES, 2007) & \href{https://assets.publishing.service.gov.uk/media/5a7aa7b6e5274a34770e630c/Letters_and_Sounds_-_DFES-00281-2007.pdf}{\nolinkurl{assets.publishing.service.gov.uk/media/5a7aa7b6e5274a34770e630c/Letters_and_Sounds_-_DFES-00281-2007.pdf}} \\
    6 & Development Matters (revised September 2023) & \href{https://assets.publishing.service.gov.uk/media/64e6002a20ae890014f26cbc/DfE_Development_Matters_Report_Sep2023.pdf}{\nolinkurl{assets.publishing.service.gov.uk/media/64e6002a20ae890014f26cbc/DfE_Development_Matters_Report_Sep2023.pdf}} \\
    7 & The reading framework (July 2023) & \href{https://assets.publishing.service.gov.uk/media/664f600c05e5fe28788fc437/The_reading_framework_.pdf}{\nolinkurl{assets.publishing.service.gov.uk/media/664f600c05e5fe28788fc437/The_reading_framework_.pdf}} \\
    \bottomrule
  \end{tabularx}
  \caption{Source documents used to build the English benchmark (United Kingdom).}
  \label{tab:sources-en}
\end{table}

\clearpage
\begin{table}[p]
  \centering
  \scriptsize
  \setlength{\tabcolsep}{3pt}
  \renewcommand{\arraystretch}{0.92}
  \resizebox{\linewidth}{!}{%
  \begin{tabular}{rlrl@{\hspace{14pt}}rlrl}
    \toprule
    \# & Model & Acc. & Company & \# & Model & Acc. & Company \\
    \midrule
    1 & Grok-4.6 & 99.27 & xAI & 61 & Gemini-3.1 Flash-Lite (Preview) & 93.43 & Google \\
    2 & Qwen3.8 Max & 99.27 & Alibaba & 62 & GPT-5.1 (Medium) & 93.43 & OpenAI \\
    3 & GLM-5.3 & 99.27 & Z.ai & 63 & GPT-5 Mini (Medium) & 93.43 & OpenAI \\
    4 & Deepseek V4 Pro & 98.54 & DeepSeek-AI & 64 & Claude Sonnet 5 & 93.43 & Anthropic \\
    5 & Gemini-3.1 Pro & 98.54 & Google & 65 & Claude Opus 5 & 92.70 & Anthropic \\
    6 & Kimi K2.6 & 98.54 & Moonshot AI & 66 & Gemini-3.5 Flash-Lite & 92.70 & Google \\
    7 & Grok-4.20 Reasoning & 98.54 & xAI & 67 & Qwen3.5 Flash & 92.70 & Alibaba \\
    8 & GPT-5.5 & 98.54 & OpenAI & 68 & GPT-5 Nano (Medium) & 92.70 & OpenAI \\
    9 & Deepseek V4 Pro (0813) & 98.54 & DeepSeek-AI & 69 & gpt-oss 120B & 92.70 & OpenAI \\
    10 & Qwen3.8 2.4T A95B & 98.54 & Alibaba & 70 & Gemini-3 Flash & 91.97 & Google \\
    11 & Qwen3.6 Max Preview & 98.54 & Alibaba & 71 & o3-Mini & 91.97 & OpenAI \\
    12 & Inkling & 98.54 & Thinking Machines & 72 & Claude Sonnet 4.6 & 91.97 & Anthropic \\
    13 & Qwen3.7 Max & 98.54 & Alibaba & 73 & Seed 1.6 Flash & 91.97 & ByteDance \\
    14 & Claude Fable 5 & 98.54 & Anthropic & 74 & Nemotron-3.5 Lightning & 91.97 & NVIDIA \\
    15 & Grok-4.5 & 98.54 & xAI & 75 & Qwen3.5 9B & 91.73 & Alibaba \\
    16 & GLM-5.1 & 98.54 & Z.ai & 76 & Muse Glimmer 30B & 91.24 & Meta \\
    17 & Qwen3.5 122B A10B & 97.81 & Alibaba & 77 & Gemma-4 31B & 91.24 & Google \\
    18 & o3 & 97.81 & OpenAI & 78 & GPT-4.1 & 91.24 & OpenAI \\
    19 & Kimi K3 & 97.81 & Moonshot AI & 79 & Grok-4.20 & 89.78 & xAI \\
    20 & Deepseek V4 Flash & 97.81 & DeepSeek-AI & 80 & GLM-4.6V & 89.78 & Z.ai \\
    21 & Gemini-3.5 Flash & 97.81 & Google & 81 & Nemotron-3 Nano & 89.78 & NVIDIA \\
    22 & GPT-5 (Medium) & 97.08 & OpenAI & 82 & Gemma-4 26B A4B & 88.32 & Google \\
    23 & Gemini-2.5 Pro & 97.08 & Google & 83 & Mistral Large 3 & 88.32 & Mistral \\
    24 & Qwen3.5 Plus & 97.08 & Alibaba & 84 & GLM-4.7 Flash & 88.06 & Z.ai \\
    25 & GPT-5.4 & 97.08 & OpenAI & 85 & GPT-5.4 Nano (Medium) & 86.86 & OpenAI \\
    26 & GPT-5.6 Terra & 97.08 & OpenAI & 86 & GPT-5.4 Mini & 84.67 & OpenAI \\
    27 & GPT-5.6 Sol & 97.08 & OpenAI & 87 & Deepseek V3 0324 & 84.67 & DeepSeek-AI \\
    28 & Inkling Small & 97.08 & Thinking Machines & 88 & GPT-4.1 Mini & 83.94 & OpenAI \\
    29 & GPT-5.4 (Medium) & 97.08 & OpenAI & 89 & Sonar & 83.21 & Perplexity \\
    30 & Qwen3.6 Plus & 97.08 & Alibaba & 90 & Qwen-3 235B A22B & 82.48 & Alibaba \\
    31 & Grok-4.3 & 97.08 & xAI & 91 & Llama-4 Maverick & 81.02 & Meta \\
    32 & Kimi K2.5 & 97.08 & Moonshot AI & 92 & Mistral Medium 3.5 & 79.56 & Mistral \\
    33 & Qwen3.5 397B A17B & 96.95 & Alibaba & 93 & Mistral Medium 3 & 77.37 & Mistral \\
    34 & Qwen3.6 35B A3B & 96.35 & Alibaba & 94 & Mistral Medium 3.1 & 77.37 & Mistral \\
    35 & Qwen3.7 Flash & 96.35 & Alibaba & 95 & Claude Haiku 4.5 & 76.64 & Anthropic \\
    36 & Gemini-3.6 Flash & 96.35 & Google & 96 & Gemma-4 12B & 75.91 & Google \\
    37 & Muse Spark 1.2 & 96.35 & Meta & 97 & Mistral Small 4 & 74.45 & Mistral \\
    38 & Claude Opus 4.6 & 96.35 & Anthropic & 98 & Gemma-4 E2B & 72.99 & Google \\
    39 & Qwen3.8 27B & 96.32 & Alibaba & 99 & Llama-4 Scout & 70.07 & Meta \\
    40 & Qwen3.6 27B & 96.30 & Alibaba & 100 & Command-R+ August & 65.69 & Cohere \\
    41 & Claude Opus 4.5 & 95.62 & Anthropic & 101 & Command A & 65.69 & Cohere \\
    42 & Claude Opus 4.8 & 95.62 & Anthropic & 102 & GPT-4.1 Nano & 64.96 & OpenAI \\
    43 & o4-Mini & 95.62 & OpenAI & 103 & Phi-4 & 63.24 & Microsoft \\
    44 & Claude Opus 4.7 & 95.62 & Anthropic & 104 & Gemma-3 27B & 62.04 & Google \\
    45 & Qwen3.5 27B & 95.62 & Alibaba & 105 & Gemma-3 12B & 61.31 & Google \\
    46 & Claude Opus 4.7 Adaptive Thinking & 95.62 & Anthropic & 106 & Mistral Small 3 & 58.39 & Mistral \\
    47 & Gemini-3.7 Flash & 95.62 & Google & 107 & Gemma-2 27B & 52.99 & Google \\
    48 & Seed 1.6 & 95.59 & ByteDance & 108 & Nova 2 Lite & 52.55 & Amazon \\
    49 & GPT-5.6 Luna & 94.89 & OpenAI & 109 & GPT-5.4 Nano & 51.82 & OpenAI \\
    50 & MiMo-V2.5 Pro & 94.89 & Xiaomi & 110 & Qwen-3 1.7B & 48.91 & Alibaba \\
    51 & GPT-5.2 (Medium) & 94.89 & OpenAI & 111 & Mistral Nemo & 44.53 & Mistral \\
    52 & MiMo-V2.5 & 94.89 & Xiaomi & 112 & Command R7B & 43.80 & Cohere \\
    53 & Qwen3.6 Flash & 94.89 & Alibaba & 113 & Qwen-3 4B & 43.80 & Alibaba \\
    54 & Qwen-3 235B A22B Thinking & 94.85 & Alibaba & 114 & Ministral 3 8B & 42.34 & Mistral \\
    55 & Gemini-2.5 Flash & 94.85 & Google & 115 & Ministral 3 3B & 41.61 & Mistral \\
    56 & Qwen3.5 35B A3B & 94.78 & Alibaba & 116 & Gemma-3 4B & 37.23 & Google \\
    57 & Claude Sonnet 4.5 & 94.16 & Anthropic & 117 & Llama-3.2 3B & 33.09 & Meta \\
    58 & MiniMax-M2.7 & 94.16 & MiniMax & 118 & Qwen-3 0.6B & 29.20 & Alibaba \\
    59 & GPT-5.4 Mini (Medium) & 94.16 & OpenAI & 119 & Llama-3.2 1B & 27.01 & Meta \\
    60 & Gemini-3.1 Flash-Lite & 94.16 & Google & & & & \\
    \bottomrule
  \end{tabular}}
  \caption{Model accuracy (\%) on the English Linguistic Knowledge Benchmark for all 119 models evaluated, sorted from highest (top left) to lowest (bottom right). Accuracy is the percentage of the 137 items answered correctly, computed over scored rows only.}
  \label{tab:all-en}
\end{table}

\begin{table}[p]
  \centering
  \scriptsize
  \setlength{\tabcolsep}{3pt}
  \renewcommand{\arraystretch}{0.92}
  \resizebox{\linewidth}{!}{%
  \begin{tabular}{rlrl@{\hspace{14pt}}rlrl}
    \toprule
    \# & Model & Acc. & Company & \# & Model & Acc. & Company \\
    \midrule
    1 & GPT-5.4 (Medium) & 97.78 & OpenAI & 62 & o4-Mini & 80.00 & OpenAI \\
    2 & Gemini-3.7 Flash & 97.04 & Google & 63 & Llama-4 Maverick & 80.00 & Meta \\
    3 & Gemini-3 Flash & 96.30 & Google & 64 & Sonar & 80.00 & Perplexity \\
    4 & Gemini-3.1 Pro & 95.56 & Google & 65 & GPT-5.4 Mini & 80.00 & OpenAI \\
    5 & Gemini-3.5 Flash & 95.56 & Google & 66 & MiniMax-M3 & 79.55 & MiniMax \\
    6 & Deepseek V4 Pro & 95.56 & DeepSeek-AI & 67 & Muse Glimmer 30B & 79.26 & Meta \\
    7 & Claude Opus 4.7 Adaptive Thinking & 95.56 & Anthropic & 68 & MiMo-V2.5 & 77.78 & Xiaomi \\
    8 & GPT-5.6 Sol & 95.56 & OpenAI & 69 & Gemma-4 26B A4B & 77.78 & Google \\
    9 & GPT-5.5 & 95.56 & OpenAI & 70 & MiniMax-M2.7 & 76.30 & MiniMax \\
    10 & Claude Opus 4.7 & 95.56 & Anthropic & 71 & Mistral Large 3 & 76.30 & Mistral \\
    11 & Claude Fable 5 & 95.56 & Anthropic & 72 & Qwen3.5 27B & 74.81 & Alibaba \\
    12 & Grok-4.5 & 95.56 & xAI & 73 & o3-Mini & 74.07 & OpenAI \\
    13 & Deepseek V4 Flash & 95.49 & DeepSeek-AI & 74 & GPT-5 Nano (Medium) & 74.07 & OpenAI \\
    14 & Claude Opus 4.8 & 94.81 & Anthropic & 75 & Qwen3.5 Flash & 74.07 & Alibaba \\
    15 & Deepseek V4 Pro (0813) & 94.81 & DeepSeek-AI & 76 & Qwen-3 235B A22B Thinking & 73.88 & Alibaba \\
    16 & Grok-4.6 & 94.07 & xAI & 77 & Qwen3.6 35B A3B & 73.13 & Alibaba \\
    17 & Kimi K3 & 94.07 & Moonshot AI & 78 & Deepseek V3 0324 & 72.59 & DeepSeek-AI \\
    18 & GPT-5.6 Terra & 93.33 & OpenAI & 79 & gpt-oss 120B & 71.85 & OpenAI \\
    19 & Claude Opus 4.6 & 92.59 & Anthropic & 80 & Qwen3.5 35B A3B & 71.64 & Alibaba \\
    20 & o3 & 92.59 & OpenAI & 81 & Qwen3.7 Flash & 69.63 & Alibaba \\
    21 & Claude Opus 5 & 92.59 & Anthropic & 82 & GPT-4.1 Mini & 68.15 & OpenAI \\
    22 & Grok-4.20 Reasoning & 92.59 & xAI & 83 & Qwen3.8 27B & 67.91 & Alibaba \\
    23 & Inkling & 92.59 & Thinking Machines & 84 & Qwen3.6 27B & 66.92 & Alibaba \\
    24 & GPT-5 (Medium) & 91.85 & OpenAI & 85 & Claude Haiku 4.5 & 65.93 & Anthropic \\
    25 & Gemini-2.5 Pro & 91.85 & Google & 86 & Gemma-4 12B & 65.67 & Google \\
    26 & Muse Spark 1.2 & 91.85 & Meta & 87 & Seed 1.6 & 63.91 & ByteDance \\
    27 & Qwen3.6 Max Preview & 91.85 & Alibaba & 88 & Mistral Medium 3 & 63.70 & Mistral \\
    28 & GPT-5.6 Luna & 91.11 & OpenAI & 89 & Mistral Medium 3.1 & 62.22 & Mistral \\
    29 & Claude Sonnet 5 & 91.11 & Anthropic & 90 & Qwen3.6 Flash & 62.22 & Alibaba \\
    30 & GPT-5.4 Mini (Medium) & 91.11 & OpenAI & 91 & Mistral Small 4 & 61.48 & Mistral \\
    31 & Gemini-3.1 Flash-Lite (Preview) & 91.11 & Google & 92 & Mistral Medium 3.5 & 60.00 & Mistral \\
    32 & Gemini-3.6 Flash & 91.11 & Google & 93 & GLM-4.6V & 59.26 & Z.ai \\
    33 & Grok-4.3 & 91.11 & xAI & 94 & GPT-5.4 Nano & 56.30 & OpenAI \\
    34 & Claude Opus 4.5 & 91.11 & Anthropic & 95 & Qwen-3 235B A22B & 55.56 & Alibaba \\
    35 & GPT-5.2 (Medium) & 91.11 & OpenAI & 96 & Gemma-3 27B & 54.81 & Google \\
    36 & Gemini-2.5 Flash & 90.37 & Google & 97 & GPT-4.1 Nano & 54.07 & OpenAI \\
    37 & Claude Sonnet 4.6 & 90.37 & Anthropic & 98 & Llama-4 Scout & 54.07 & Meta \\
    38 & Kimi K2.6 & 90.37 & Moonshot AI & 99 & Qwen3.5 9B & 51.85 & Alibaba \\
    39 & Gemini-3.5 Flash-Lite & 90.37 & Google & 100 & Command A & 50.37 & Cohere \\
    40 & GPT-5.1 (Medium) & 89.63 & OpenAI & 101 & GLM-4.7 Flash & 46.56 & Z.ai \\
    41 & Qwen3.7 Max & 89.63 & Alibaba & 102 & Gemma-2 27B & 45.93 & Google \\
    42 & Inkling Small & 88.89 & Thinking Machines & 103 & Command-R+ August & 42.22 & Cohere \\
    43 & GPT-5.4 & 88.89 & OpenAI & 104 & Gemma-4 E2B & 40.74 & Google \\
    44 & Qwen3.5 397B A17B & 88.55 & Alibaba & 105 & Nemotron-3.5 Lightning & 40.30 & NVIDIA \\
    45 & Qwen3.8 2.4T A95B & 88.15 & Alibaba & 106 & Mistral Small 3 & 40.00 & Mistral \\
    46 & Kimi K2.5 & 87.41 & Moonshot AI & 107 & Gemma-3 12B & 39.26 & Google \\
    47 & Gemini-3.1 Flash-Lite & 87.41 & Google & 108 & Nemotron-3 Nano & 38.93 & NVIDIA \\
    48 & Qwen3.8 Max & 87.41 & Alibaba & 109 & Nova 2 Lite & 35.56 & Amazon \\
    49 & Qwen3.5 Plus & 86.67 & Alibaba & 110 & Seed 1.6 Flash & 33.83 & ByteDance \\
    50 & MiMo-V2.5 Pro & 86.67 & Xiaomi & 111 & Mistral Nemo & 29.63 & Mistral \\
    51 & GLM-5.1 & 85.93 & Z.ai & 112 & Phi-4 & 28.89 & Microsoft \\
    52 & Qwen3.6 Plus & 85.93 & Alibaba & 113 & Gemma-3 4B & 26.67 & Google \\
    53 & GLM-5.2 & 85.82 & Z.ai & 114 & Ministral 3 3B & 23.70 & Mistral \\
    54 & GLM-5.3 & 85.19 & Z.ai & 115 & Ministral 3 8B & 22.96 & Mistral \\
    55 & Gemma-4 31B & 85.19 & Google & 116 & Command R7B & 22.22 & Cohere \\
    56 & Grok-4.20 & 85.19 & xAI & 117 & Llama-3.2 3B & 21.48 & Meta \\
    57 & Claude Sonnet 4.5 & 81.48 & Anthropic & 118 & Qwen-3 4B & 20.74 & Alibaba \\
    58 & GPT-5 Mini (Medium) & 80.74 & OpenAI & 119 & Llama-3.2 1B & 14.07 & Meta \\
    59 & Qwen3.5 122B A10B & 80.74 & Alibaba & 120 & Qwen-3 1.7B & 11.11 & Alibaba \\
    60 & GPT-4.1 & 80.74 & OpenAI & 121 & Qwen-3 0.6B & 10.37 & Alibaba \\
    61 & GPT-5.4 Nano (Medium) & 80.00 & OpenAI & & & & \\
    \bottomrule
  \end{tabular}}
  \caption{Model accuracy (\%) on the Kiswahili Linguistic Knowledge Benchmark for all 121 models evaluated, sorted from highest (top left) to lowest (bottom right). Accuracy is the percentage of the 135 items answered correctly, computed over scored rows only.}
  \label{tab:all-sw}
\end{table}

\clearpage
\section{Supplementary Figures}
\label{app:figures}

\begin{figure}[h]
  \centering
  \includegraphics[width=\linewidth]{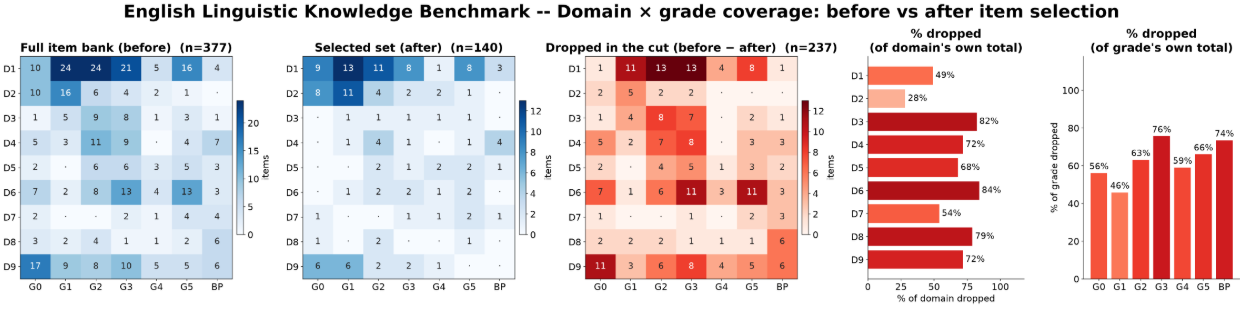}
  \caption{English item coverage across linguistic domains (D1--D9) and grade levels (G0--G5, BP = beyond primary) before and after item selection.}
  \label{fig:coverage-en}
\end{figure}

\begin{figure}[h]
  \centering
  \includegraphics[width=\linewidth]{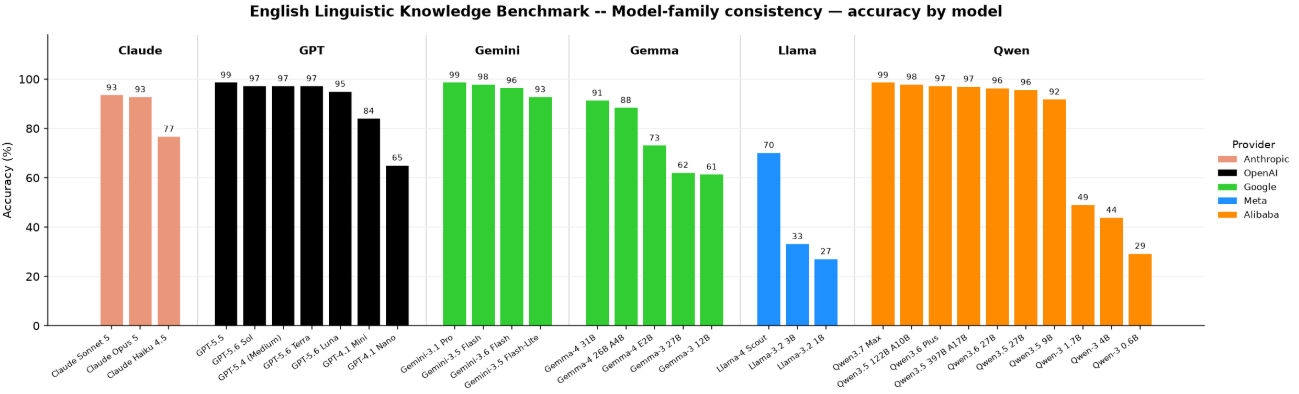}
  \caption{English Linguistic Knowledge accuracy by model, grouped by model family. Bars within each family are ordered by measured accuracy.}
  \label{fig:family-en}
\end{figure}

\begin{figure}[h]
  \centering
  \includegraphics[width=\linewidth]{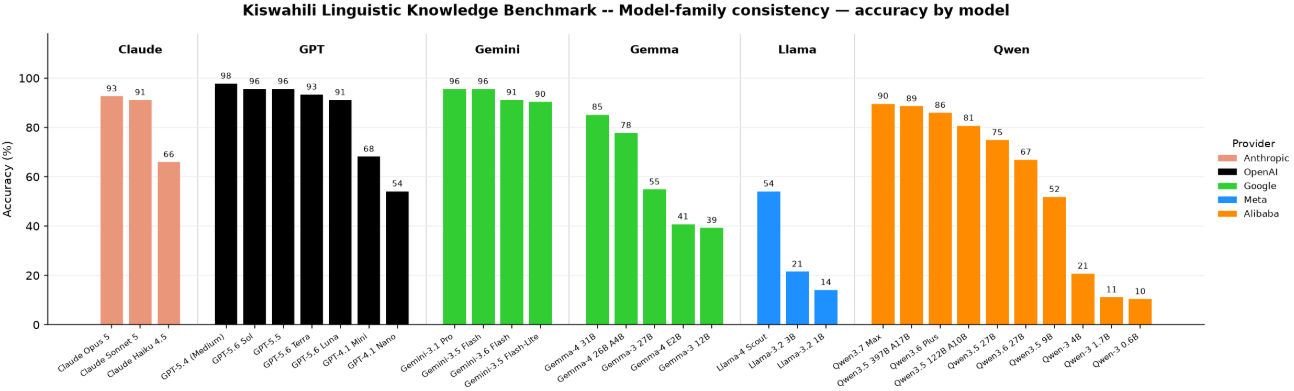}
  \caption{Kiswahili Linguistic Knowledge accuracy by model, grouped by model family. Bars within each family are ordered by measured accuracy.}
  \label{fig:family-sw}
\end{figure}

\clearpage
\section{Hardware Tier Assignment}
\label{app:tiers}

We grouped the models by the smallest class of device on which they can run locally, so that accuracy can be compared within each deployment class. For each model, we use the total number of parameters as a proxy for the memory needed at inference. We did not run any model on the devices described here, so all the requirements given below are estimates. The tiers describe deployability at an acceptable quality level, not only whether the weights can fit in memory. Two assumptions limit this further. First, we assume that the model is the main task running in the foreground and that other applications are closed; the tiers therefore do not describe normal use with several applications open at the same time. Second, the tier does not correspond to the configuration we benchmarked: all evaluations were run at the best precision available for each model, while the tiers assume a quantised deployment.

\paragraph{How requirements were estimated.}
The assignment uses three gates, applied in order. The first gate is \emph{deployment access}. We resolve it before any memory calculation, because whether a user can obtain the weights is a different question from whether their hardware can hold them. Closed-weight models are therefore assigned to the cloud tier whatever their size, because they are served only through their vendor's API. The second gate is \emph{data sufficiency}: open-weight models whose parameter count is not published are excluded from the comparison rather than assigned to a default tier. The third gate is \emph{hardware capacity}, described below.

We estimate the memory required as the total number of parameters, multiplied by the number of bytes per parameter at the precision assumed for that tier, plus 25\%. This allowance covers runtime overhead, activations, and a key-value cache for a short or medium context. The key-value cache grows with context length and the number of layers, not with the number of parameters, so the allowance is tight for the smallest models and generous for the largest.

The memory available to the model is only a fraction of the device's total RAM, and this fraction differs by tier, depending on whether memory is shared with an operating system or dedicated to an accelerator. On mobile and tablet hardware we allow the model 50\% of the total, which is a maximum for a well-behaved application rather than a typical value. At workstation level the model is placed in the GPU's dedicated memory, so we allow 85\%.

Precision is defined per tier. On mobile and tablet we assume 4 bits, because memory is limited on these devices and 4-bit quantisation is the usual way of deploying a model on them. On a workstation, memory is no longer the limiting factor and precision becomes a real choice, so we assume 8 bits and treat this as a minimum quality level. In the cloud, memory is less of a constraint, so precision never changes the tier a model is assigned to.

Storage is not used in the assignment. With each tier's quantisation, the model's size on disk stays well below the free space available on that class of device: 1.6\,GB at the top of the mobile tier, 4.8\,GB at the top of the tablet tier, and 21.8\,GB at the top of the workstation tier.

\paragraph{Tier boundaries.}
We define four tiers, set out in Table~\ref{tab:tiers}. The RAM values are device totals as they appear on a specification sheet: 4\,GB corresponds to an entry-level Android phone, 12\,GB to a premium phone or a tablet, and 32\,GB to the memory of a high-end consumer graphics card. The usable share is the part of this total the model can use once the operating system and other processes have taken what they need, and the model budget is the product of the two. The maximum number of parameters is the largest model this budget can hold at the tier's precision, after applying the 25\% allowance. Since memory is the only real constraint, the assignment reduces to a threshold on parameter count. In Figure~\ref{fig:tiers}, the mobile and tablet tiers are presented together.

\begin{table}[h]
  \centering
  \footnotesize
  \setlength{\tabcolsep}{4pt}
  \begin{tabular}{lrrrrr>{\raggedright\arraybackslash}p{3.4cm}}
    \toprule
    Tier & Device RAM & Usable share & Model budget & Precision & Max.\ params & Indicative device \\
    \midrule
    Mobile & 4\,GB & 50\% & 2.0\,GB & 4-bit & 3.2B & Entry-level Android, edge device \\
    Tablet & 12\,GB & 50\% & 6.0\,GB & 4-bit & 9.6B & Premium phone, tablet \\
    Workstation & 32\,GB & 85\% & 27.2\,GB & 8-bit & 21.8B & Single high-end consumer GPU \\
    Cloud & --- & --- & unbounded & 8/16-bit & above & Datacentre GPU, multi-GPU server \\
    \bottomrule
  \end{tabular}
  \caption{Hardware tiers used to group models by deployment class.}
  \label{tab:tiers}
\end{table}

\paragraph{Special cases.}
Sparse (mixture-of-experts) models receive no reduction for the smaller number of parameters they activate per token. The router can choose any expert for any token, so the complete model has to stay in memory, and the total parameter count therefore determines memory exactly as for a dense model. We initially assumed the opposite, and gpt-oss-120b showed that this assumption was wrong in practice. This model has 117 billion parameters in total and activates 5.1 billion per token, so it would belong to the tablet tier if only the second value were considered. Its published requirement, however, is within 80\,GB of memory, and local deployments reported by users need about 64\,GB. Our estimate at 4 bits with the 25\% allowance gives 73\,GB, close to the published value. This agreement is partly a coincidence: the allowance is least appropriate for a model of this size, and the model's native format is slightly above 4 bits, so the two errors partly compensate each other.

Finally, the cloud tier contains two different kinds of model: those whose weights cannot be obtained, and those too large for consumer hardware. Membership of the cloud tier should therefore not be interpreted as meaning that a model is large.

\end{document}